# Progress and summary of reinforcement learning on energy management of MPS-EV

Jincheng Hu[1+], Yang Lin[2+], Liang Chu[2], Zhuoran Hou[2], Jihan Li[1], Jingjing Jiang[1], Yuanjian Zhang[1*]

*1, Department of Aeronautical and Automotive Engineering, Loughborough University, Leicestershire LE11 3TU, UK*

*2, College of Automotive Engineering, Jilin University, 5988 Renmin Street, Changchun 130022, China*

**Abstract**

The high emission and low energy efficiency caused by internal combustion engines (ICE) have become unacceptable under environmental regulations and the energy crisis. As a promising alternative solution, multi-power source electric vehicles (MPS-EVs) introduce different clean energy systems to improve powertrain efficiency. The energy management strategy (EMS) is a critical technology for MPS-EVs to maximize efficiency, fuel economy, and range. Reinforcement learning (RL) has become an effective methodology for the development of EMS. RL has received continuous attention and research, but there is still a lack of systematic analysis of the design elements of RL-based EMS. To this end, this paper presents an in-depth analysis of the current research on RL-based EMS (RL-EMS) and summarizes the design elements of RL-based EMS. This paper first summarizes the previous applications of RL in EMS from five aspects: algorithm, perception scheme, decision scheme, reward function, and innovative training method. The contribution of advanced algorithms to the training effect is shown, the perception and control schemes in the literature are analyzed in detail, different reward function settings are classified, and innovative training methods with their roles are elaborated. Finally, by comparing the development routes of RL and RL-EMS, this paper identifies the gap between advanced RL solutions and existing RL-EMS. Finally, this paper suggests potential development directions for implementing advanced artificial intelligence (AI) solutions in EMS.

**Keywords:** Reinforcement learning, multi-power source electric vehicles, energy management strategy.

## 1. OVERVIEW

The restriction of environmental regulations and the rise of clean energy technologies have accelerated electric vehicles replacing conventional vehicles [1]. Compared with ICE vehicles, battery-powered electric vehicles (BEVs) have been proven to have significant advantages in emission, economy, and efficiency. However, limited battery capacity causes mileage anxiety [2]. Multi-power source electric vehicles (MPS-EVs)

[+] These authors contribute equally to this paper and should be considered as co-first author

* Corresponding author. Tel.: (44) 07422554692

*E-mail address:* author@institute.xxx .

provide two or more power sources that can work together. The engine and motor are used for driving. The fuel tank, hydrogen tank, and other devices are used for storage. By regulating the output power from multiple sources, their advantages can be better exploited. It is necessary for MPS-EVs to adjust the working mode according to the changeable working conditions. This requires the vehicle to have a comprehensive EMS to determine the power distribution. An appropriate and effective EMS can significantly improve energy efficiency, and achieve lower emissions and higher economy. [3].

Rule-based and optimization-based methods are the two main categories of EMS [4]. At present, rule-based methods are the most widely used in the real world. The reason why they can be applied to practice is their robustness. Created by engineers with a wealth of expert experience, rule-based can adapt to various driving cycles without the risk of battery wearing out. They are easy to develop and safe to apply [5]. Simple rules can make it easier for the controller to operate. The representative algorithms are fuzzy logic (FL) mode [6], charge consumption charge maintenance (CD-CS) mode [7], and so on. Although it is easy to be applied in the real world, the improvement potential of the rule-based methods in fuel economy is usually lower than that of optimization-based methods. By setting the cost function in advance, the researchers train the EMS to make control signals matching with the working condition to optimize some indexes and reduce the corresponding cost, such as improving fuel economy, prolonging the service life of components, and reducing emissions. This enables the optimization-based methods to perform better on specific driving cycles. Optimization-based algorithms can be further divided into global optimization and real-time optimization [8]. Global optimization needs to obtain the whole driving cycle in advance and calculate the global optimal solution based on the predefined speed profiles of the entire process. One of the representatives of global optimization is dynamic programming (DP). As a mathematical optimization method and computer programming method, DP divides the complex calculation process into solving subproblems [9] and gradually obtains the global optimal solution in a recursive way. Some researchers used DP to develop EMS and significantly improved the control effect [10, 11]. However, it is impossible to obtain accurate speed information in advance and bear the high computational expense for driving cycles lasting thousands of seconds. DP cannot be applied in real time. In more cases, it is used as a benchmark to judge the performance of other methods. Some researchers also apply the idea of DP to inspire rule-based or real-time optimization methods [4, 12].

EMS based on a real-time optimization algorithm emphasizes real-time performance and outputs a control signal to minimize the cost with the cost at the current time as the objective function [13]. This method reduces the computational cost of the control strategy and improves the real-time performance, making it possible to apply the optimization-based method to the real road. Typical representatives are equivalent consumption minimization strategy [14] (ECMS) and model predictive control [15] (MPC). ECMS constructs the final cost by superimposing the consumption of different types of energy at the current moment by setting an equivalence factor. It takes minimizing the equal fuel consumption cost at the current moment as the objective of the control strategy [16,17]. However, the equivalence factor needs to be specified artificially, an inappropriate value may reduce the fuel economy and bring the risk of state of charge (SOC) for the battery exceeding the constraints, which results in the reliance on expertise. Similar to ECMS, the MPC controller calculates a future control sequence in which all control signals meet the objective of minimizing energy costs. It then takes the first action of the calculated control sequence as the executed control signal. This process is repeated at the next time step by moving the prediction range forward one step [15]. Some researchers use MPC to develop EMS based on accurate models [18-20]. Although the real-time optimization algorithm solves the problem of real-time application in the driving process, there is still a large gap between the optimization performance of the whole driving cycle and global optimization because of its local optimization characteristics.

Reinforcement learning as an artificial intelligence method with excellent real-time performance and global optimization-like capabilities is expected to solve the MPS-EV energy management optimization problem [21]. In recent years, reinforcement learning has achieved beyond human-level results on a range of domains such as Go [22] and Google Research Football [23], which demonstrates its potential for application to complex control problems. As a highly concerned artificial intelligence method, RL brings a new solution to the development of MPS-EV EMS. RL models the control problem. During the training process of the algorithm, the agent interacts with the environment and improves the control strategy based on the results of the interaction, through which the RL algorithm can achieve self-learning and gradually approach the global optimal solution [24]. Compared with the rule-based method, RL is based on optimization theory and can achieve better control effects; compared with global optimization algorithms such as DP, RL can be applied in real-time and has lower computational cost, which makes it have better application potential; compared with real-time optimization algorithms such as ECMS and MPC, reinforcement learning has self-learning characteristics [25], requires less prior knowledge and model construction, and the control effects is closer to global optimization [21]; these features have attracted wide attention from researchers in academia and industry, and more and more studies are applying reinforcement learning to MPS-EV energy management problems.

This paper provides an in-depth generalization and summary of the application of RL methods in MPS-EV energy management, and analyzes the current status of RL-EMS research in the following five aspects:

1) Algorithmic applications: Reinforcement learning algorithms are a family of artificial intelligence algorithms that originated from behaviorism and have been further extended by the rise of deep learning in the last decade. Through an extensive literature study and survey, this paper summarizes the types of RL algorithms applied in MPS-EV energy management problems in general and summarizes the applications of the algorithms based on chronological order.
2) State Sensing: In the energy management problem, the state describes the sensing capability of the EMS. It specifies the ability and scope of the EMS to acquire information and is the sole basis for algorithmic decisions. However, the definition of state variables varies in different EMS studies. Therefore, this paper summarizes the definition and design of the state of reinforcement learning solution variables and classifies them into two categories: in-vehicle state variables and out-of-vehicle state variables for a comprehensive introduction.
3) Decision schemes. The introduction of reinforcement learning has overturned the traditional EMS research approach and enriched the design of control schemes for vehicle power signals. In this paper, we list three control frameworks for final energy distribution and classify them into end-to-end control, hierarchical control, and parallel control for in-depth discussion.
4) Reward setting. As a self-learning method, the reward function of reinforcement learning shows the iterative goal of the algorithm. This paper summarizes a series of metrics commonly found in RL-EMS such as equivalent fuel consumption and component maintenance and their expressions in the function.
5) Training methods: The selection of training data and hyperparameter tuning have a great impact on the final control effect of RL solutions. In this paper, we analyze RL-EMS studies in terms of training data diversity, parameter initialization, and other improvements in training methods.

The rest of this paper is organized as follows: Section 2 introduces the popular types of multi-energy source vehicles, the general solution process and optimization objectives of RL-EMS research. Section 3 reviews the recent applications of reinforcement learning algorithms to energy management strategies from five directions, and introduces the progress of each research and improvements to traditional RL-EMS. Section 4 provides an overview of RL-EMS and an outlook on the current challenges and future research directions of RL-EMS.

## 2. Reinforcement Learning and Energy Management of MPS -EV

### *2.1. Energy management of MPS-EV*

#### *2.1.1. Representations of MPS-EV Configuration*

With the maturity of vehicle electrification, electrified powertrains have allowed for high-power electrical transmission, making it possible for electric motors to do work instead of engines. While the power performance is improved, the development of battery technology provided more adequate energy reserves with far less pollution than fossil fuels [26]. Relevant technologies have helped to develop a series of multi-power source electric vehicle configurations. Different types of MPS-EVs are equipped with various energy storage devices to make use of clean energy by utilizing electricity or hydrogen to keep driving. Hybrid electric vehicles (HEVs) and plug-in hybrid electric vehicles (PHEVs) take batteries and fuel tanks as energy storage devices, while fuel cell vehicles (FCEVs) are equipped with fuel cells, batteries, or supercapacitors.

A) Hybrid Electric Vehicle & Plug-in Hybrid Electric Vehicle

Relying on the driving force generated by the motor and the engine, the mechanical power required to drive HEVs and PHEVs comes directly or indirectly from the engine to some extent. The retention of the conventional internal combustion engine in the electrified powertrain inevitably brings fossil fuel emissions and a complex structure [27]. However, in contrast to conventional engine-powered vehicles, the hybrid powertrain adjusts the engine working points through the conversion between electrical and mechanical energy. On the one hand, under the conditions of low engine efficiency, such as idling and low load, the hybrid drive system can completely use the motor instead of the engine to do work and consume the electric power of the battery; on the other hand, it can also increase the engine output power until the working point falls in the high-efficiency zone. The part beyond the demanded power is stored in the battery [28] to provide power for possible discharges in the future. As shown in Fig. 1, the commonly used configurations in HEV & PHEV can be divided into series hybrid, parallel hybrid, and series-parallel hybrid according to the connection mode of different drive components [29]. Different configurations usually correspond to different mixing degrees, which is the proportion of the motor output power in the whole system output power. The series hybrid configuration is the closest to the pure electric vehicle, and the parallel hybrid has the best dynamic performance.

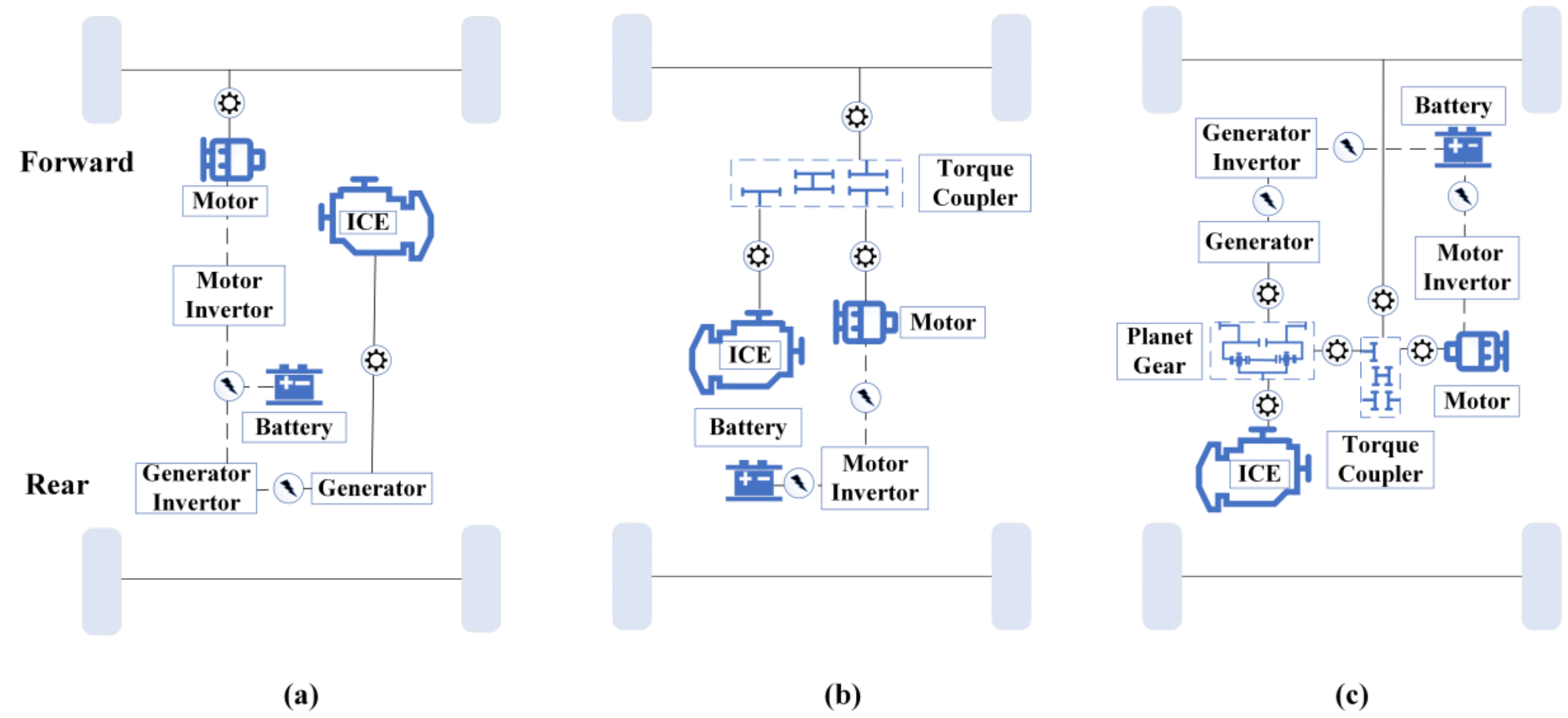

**Fig. 1.** Powertrain architectures of HEV&PHEV, (a)series hybrid, (b)parallel hybrid, (c)series-parallel hybrid.

B) Fuel Cell Electric Vehicle

FCEVs rely on electricity the onboard fuel cell generates to keep moving. The fuel cell (FC) is a chemical device that converts the chemical energy possessed by the fuel directly into electrical energy, and the widely used fuel is high-purity hydrogen. Fig. 2 shows two main configurations of the FCEV powertrain: the fuel cell/battery (FC/B) configuration and the fuel cell/battery/supercapacitor (FC/SC/B) configuration. The current fuel cell widely used in FCEV is the polymer electrolyte membrane fuel cell (PEMFC), which has a high conversion efficiency in the electrochemical process [30]. Hydrogen reacts with oxygen in the fuel cell to produce electrical energy to drive the motor, which is mechanically connected to the drive wheel. Because of the complex and variable driving cycles, the powertrain requires an unstable power output. While the fuel cell is suitable for operation in a steady state to prolong its service life [31], the battery is employed to perform the corresponding charge and discharge operations to adjust the fuel cell output power. FCEVs that take hydrogen as fuel are faster to refuel and have better economy and fuel efficiency [32]. In addition to batteries, some FCEVs incorporate supercapacitors (SCs) as energy storage devices. The simple structure of SCs and high charging and discharging efficiency allow them to respond quickly to the changing demand power [33].

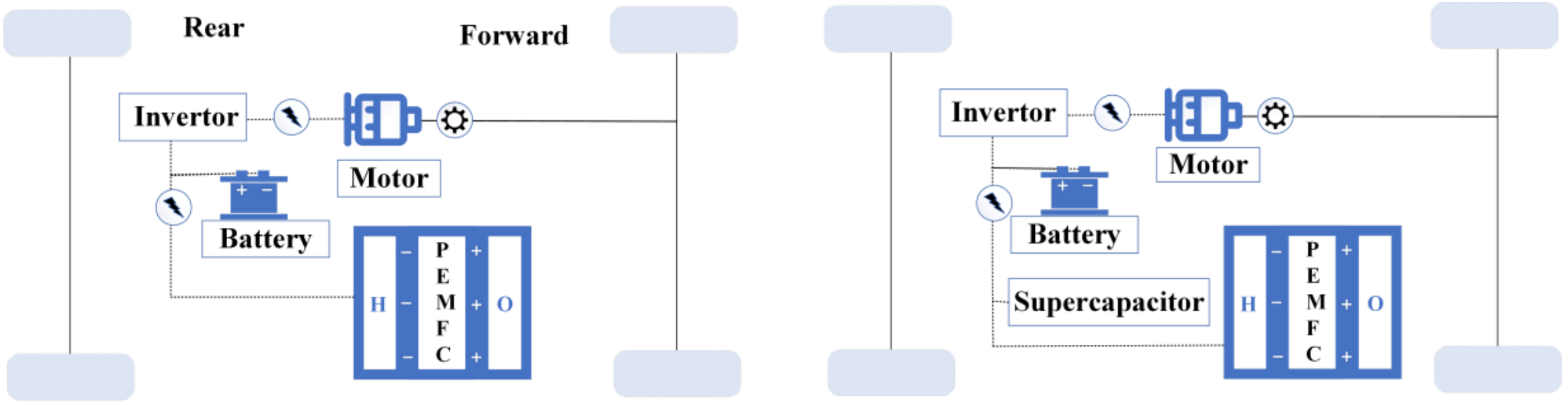

(a) (b)

**Fig. 2.** Powertrain architectures of FCEV, (a) FC + B, (b) FC + SC + B.

### *2.1.2. Modeling of energy management problem of MPS-EV*

In the real-world driving process, the driver changes the pedal position to control the vehicle speed. In the simulation, the acceleration and deceleration process of the vehicle model is simulated based on the speed profile provided by the predefined training driving cycles, and the power demand of the multi-energy source vehicle is then determined. The MPS-EV energy management problem is dedicated to the rational distribution of energy flow under different working conditions to improve fuel economy. The consumption of electricity and fuel is the most intuitive reflection of the economy of the driving process and is also a standard index to evaluate the control effect of the EMS. Some research takes into account the effect of energy distribution on the health condition of components, such as the adverse effects of frequent start-stop on engine and battery life [34,35]; fuel consumption and battery SOC are directly affected by control actions by energy distribution, and a good EMS can finish the testing driving cycle with less power and fuel consumption by balancing the distribution of multi-power sources.

Taking the backward model [36, 37, 38] as an example, Fig. 3 illustrates the general process of the energy management problem: 1) working condition requirements: the driving cycle of the vehicle is provided based on the training data. The speed profile of the driving cycle and vehicle parameters such as vehicle mass and windward area determine the vehicle output power during the simulation; 2) control actions: the EMS distributes the energy flow, and the vehicle operating mode is determined by the control signals from the EMS. The engine, motor, and generator work in the given state; 3) performance analysis: the energy sources output power, with the engine, battery, and other models to calculate and record the fuel and power consumption. At the end of the driving cycle, relevant metrics would be used to evaluate the effectiveness of the EMS. Meanwhile, during the EMS solution to the energy management problem, the relevant physical quantities should be constrained: SOC, engine, and motor torque speed should all be within a certain range.

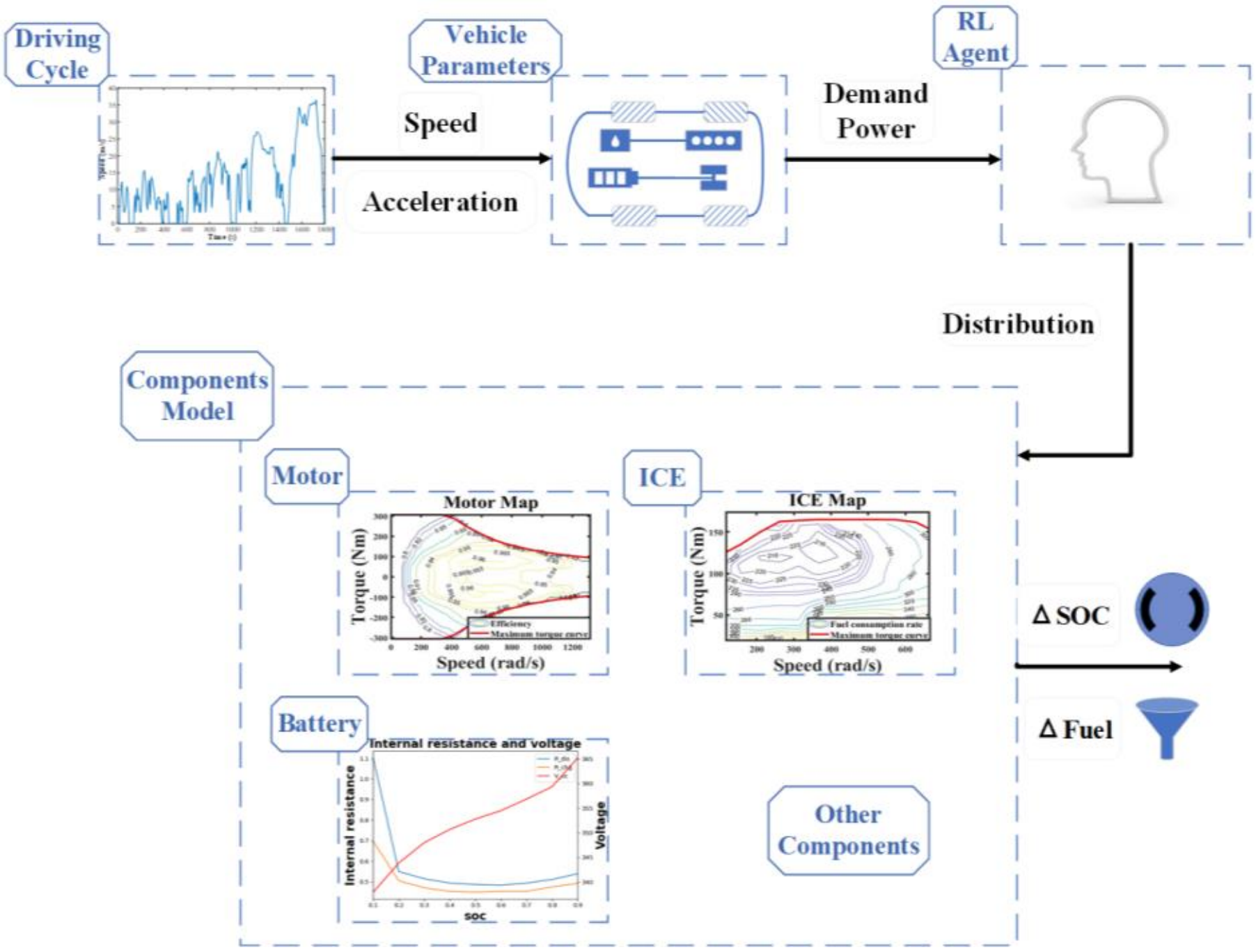

**Fig. 3.** General process of MPS-EV energy management under the control of RL.

## *2.2. Reinforcement learning and Markov decision process*

Reinforcement learning sets an agent to interact with its environment, and the agent spontaneously updates its policy for the interaction through the rewards it receives [39]. In RL-based EMS research, the agent is the controller that determines the distribution of the energy flow; the vehicle model and driving cycles are modeled as the environment in which the agent resides. The relevant metrics of the vehicle are quantified as the rewards received by the agent through a reward function. An essential feature of reinforcement learning is the self-learning ability of the agent. As shown in Fig.4, the agent observes the state of the environment and selects an action according to its current policy. The action selection results from the balance between exploration and exploitation. The agent has to try different control actions to enrich its experience. At the same time, it also has to ensure the efficiency of exploration and avoid useless and dangerous exploration. The environment is influenced by the action, returning a reward and presenting a new state [40], providing experiences for the agent to conduct policy iteration.

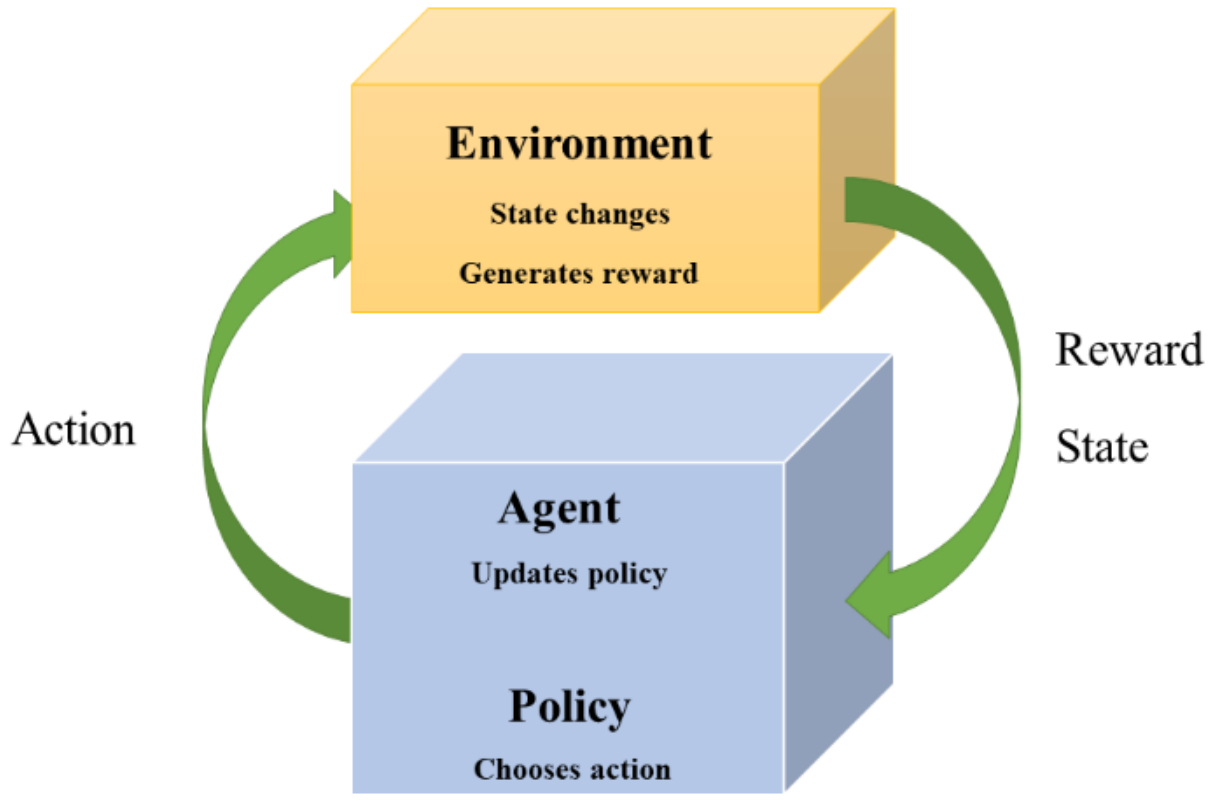

**Fig. 4.** Interaction between agent and environment.

The interaction process between the agent and environment is usually modeled as a Markov decision process (MDP), in which the state at the next step is only related to the current state and has nothing to do with the previous state [41]. At each time step, the agent receives the observation indicating vehicle condition from the environment and selects an action to determine the energy distribution of the multi-power sources. After one time step, the vehicle is affected by the action of the agent and enters a new state. The reward is calculated and returned to the agent. This process continues to cycle until the end of the driving cycles and constitutes a typical Markov decision process [42]. It is worth noting that in the current research, the Markov process of RL-EMS is completely observable. The state transition probability matrix of vehicle speed in the Markov chain of one iteration is basically unchanged and known. There is no hidden variable influence. The Markov decision process includes five elements: $(X, A, P_{XA}, \gamma, R)$ [43]. $X$ represents the state space of the Markov process, which contains all possible values of the state variables. $A$ represents the action space of the agent. $P_{XA}$ represents the state transition probability for each $X$ and action $A$, $P_{XA}$ gives the distribution of the next state that the environment will transition to after agent take action $A$ in state $X$. $\gamma$ represents the discount rate, and the value is between 0 and 1. By adjusting this value, the agent can balance the importance of the current reward and subsequent rewards. $R$ represents the reward obtained by the agent after making the action. These five elements are essential factors in the reinforcement learning training process.

## 3. RL-EMS studies for MPS-EV

The recent research in multi-power source vehicle energy management is classified and discussed in this section. From the RL framework and MDP elements introduced in the previous section, it is clear that after selecting the algorithms used, the RL-based EMS can be trained by completing the modeling of the main factors in the MDP. This section first introduces the algorithms used in the recent research according to the order of development from reinforcement learning to deep reinforcement learning, listing the advances made by various algorithms in the EMS field; secondly, the state variables perceived by the agent in the MDP are enumerated in detail; then the selection of actions for the agent and the corresponding control framework are described; the settings of the reward functions are shown categorically; finally, the novel training method based on the practical characteristics of the energy management problem is presented. Fig. 5 and Table 1 show the basic settings of the different types of algorithms in MPS-EV energy management and the recurrent driving cycles in the training data.

- Table 1. Solutions in RL-EMS for MPS-EVs.

| Algorithm | Configuration | State | Action | Reward | Reference |
|---|---|---|---|---|---|
| Q-learning | HEV (parallel) | $v$, $T_{dem}$ | $T_m$ | *fuel, elec* | [47] |
| | | $v$, $SOC$, $T_{dem}$, $G$ | $T_m$ | *fuel, elec* | [66] |
| | | $v$, $SOC$, $T_{dem}$, $G$ | $T_m$ | *fuel, elec, ΔSOC* | [50] |
| | | $v$, $SOC$, $P_{dem}$ | $I$ | *fuel* | [51] |
| | HEV&PHEV (parallel) | $v$, $SOC$, $P_{dem}$ | $I$ | *fuel, deg* | [53] |
| | HETV | $SOC$, $n_g$ | $T_{eng}$ | *fuel, ΔSOC* | [48] |
| | | $SOC$, $n_{eng}$ | $Thr_{eng}$ | *fuel, elec* | [54] |
| | | $SOC$, $n_g$ | $Thr_{eng}$ | *fuel, elec* | [55] |
| SARSA | PHEB | $v$, $SOC$, $P_{dem}$ | $P_{eng}$, $G$ | *fuel, elec* | [56] |
| | HETV | $SOC$, $n_g$ | $T_{eng}$ | *fuel, ΔSOC* | [58] |
| | FCEV (FC + Battery) | $SOC$, $ped$ | $DOH$ | *ΔSOC, deg* | [59] |
| Dyna | HEV (series) | $SOC$, $P_{dem}$ | $P_{eng}$ | *fuel, ΔSOC* | [65] |
| | HEV (parallel) | $SOC$ | $Thr_{eng}$ | *fuel, ΔSOC* | [62] |
| | PHEV (series) | $SOC$ | $mod$, $T_m$, $T_{eng}$ | *fuel, elec* | [64] |
| | HETV | $SOC$, $n_{eng}$ | $Thr_{eng}$ | *fuel, ΔSOC* | [61] |
| DQN, Dyna | HETV | $SOC$, $P_{dem}$, $n_g$ | $Thr_{eng}$ | *fuel, ΔSOC* | [63] |
| DQN | HEV (series) | $v$, $acc$, $SOC$, $P_{eng}$ | $\Delta P_{eng}$ | *fuel, elec* | [77] |
| | HEV (parallel) | $SOC$, $T_{dem}$ | $T_{eng}$ | *fuel* | [70] |
| | | $v$, $SOC$, $P_{dem}$, $dis$ | $T_{eng}$ | *fuel* | [71] |
| | PHEV (series-parallel) | $SOC$, $P_{dem}$, $dis$ | $P_{eng}$ | *fuel* | [78] |
| | PHEB | $v$, $acc$, $SOC$, $P_{eng}$ | $\Delta P_{eng}$ | *fuel, elec* | [67] |
| | HETV | $v$, $acc$, $SOC$, $P_{dem}$, $\omega$ | $\Delta n_{eng}$ | *fuel, ΔSOC* | [69] |
| | | $SOC$, $P_{dem}$, $n_g$ | $Thr_{eng}$ | *fuel, ΔSOC* | [72] |
| | | $SOC$, $P_{dem}$, $n_g$ | $Thr_{eng}$ | *fuel, ΔSOC* | [73] |
| | FCEV(FC + Battery) | $v$, $SOC$, $P_{fc}$ | $\Delta P_{fc}$ | *fuel, ΔSOC, deg* | [74] |
| DDPG | HEV (parallel) | $v$, $SOC$, $T_{dem}$, $G$ | $P_m$ | *fuel, elec* | [36] |
| | HEV (series, parallel, series-parallel) | $v$, $acc$, $SOC$ | $T_{eng}$，$n_{eng}$，$T_m$ | *fuel, ΔSOC, elec* | [37] |
| | HETV | $v$, $acc$, $SOC$ | $P_{eng}$ | *fuel, ΔSOC* | [80] |
| | | $v$, $SOC$, $P_{dem}$, $n_{eng}$ | $Thr_{eng}$ | *fuel, ΔSOC* | [81] |
| | | $v$, $acc$, $SOC$, $\omega$, $\omega'$ | $n_{eng}$, $T_{eng}$ | *fuel, ΔSOC, elec* | [82] |
| | FCEV (FC + Battery) | $v$, $acc$, $SOC$ | $P_{fc}$ | *fuel, ΔSOC* | [83] |
| TD3 | HEV (parallel) | $v$, $acc$, $SOC$, $m$, $\theta$ | $T_{eng}$ | *fuel, ΔSOC* | [85] |
| | HEV (series-parallel) | $v$, $acc$, $SOC$ | $P_{eng}$ | *fuel, ΔSOC* | [86] |
| | FCEV (FC + Battery) | $v$, $acc$, $SOC$, $P_{fc}$, $P_{dem}$, $n_p$ | $P_{fc}$ | *fuel, ΔSOC, deg* | [87] |
| | | $v$, $acc$, $SOC$, $P_{fc}$, $P_{dem}$, $n_p$ | $P_{fc}$ | *fuel, ΔSOC, deg* | [34] |
| A2C | HEV (parallel) | $v$, $acc$, $SOC$ | $\Delta P_{eng}$ | *fuel, ΔSOC* | [89] |
| | HEV (series-parallel) | $v$, $acc$, $SOC$, $P_{dem}$, $\theta$ | $mod$, $n_{eng}$, $T_{eng}$, $T_m$ | *fuel, ΔSOC* | [90] |
| | PHEB | *v, acc, SOC* | *mod*, $T_{eng}$, $n_{eng}$, $T_m$ | *fuel*, $\Delta SOC$ | [88] |
| PPO | HEV (series-parallel) | *v, acc, SOC*, $n_{eng}$ | $T_{eng}$, $n_{eng}$ | *fuel*, $\Delta SOC$ | [35] |
| | | *v, acc, SOC* | $T_{eng}$ | *fuel*, $\Delta SOC$ | [38] |
| SAC | HEB | *tem, SOC*, $v_{wheel}$, $T_{wheel}$, *vol* | $P_{eng}$ | *fuel*, $\Delta SOC$ | [96] |

Abbreviation. *v* Vehicle speed. ω Yaw rate of the tracked vehicle. *m* Load mass. *ped* Pedal position. *acc* Vehicle acceleration. *SOC* Battery state of charge. $v_{wheel}$ Driving wheel speed. $T_{wheel}$ Driving wheel torque. $T_{dem}$ Torque demand. $P_{dem}$ Power demand. *G* Gear ratio. $n_g$ Speed of generator. $n_{eng}$ Speed of engine. $Thr_{eng}$ Throttle angle of engine. $P_{eng}$ Power of engine. $T_{eng}$ Torque of engine. $T_m$ Torque of motor. $P_m$ Power of motor. $P_{fc}$ Power of fuel cell. *I* Battery current. *DOH* Degree of hybridization. *Tem* Battery temperature. *vol* Battery voltage. $n_p$ Number of passengers. *dis* Distance from destination. *mod* Work mode. *θ* Road gradient. *fuel* Fuel consumption. *elec* Electric energy consumption. *deg* FC degradation.

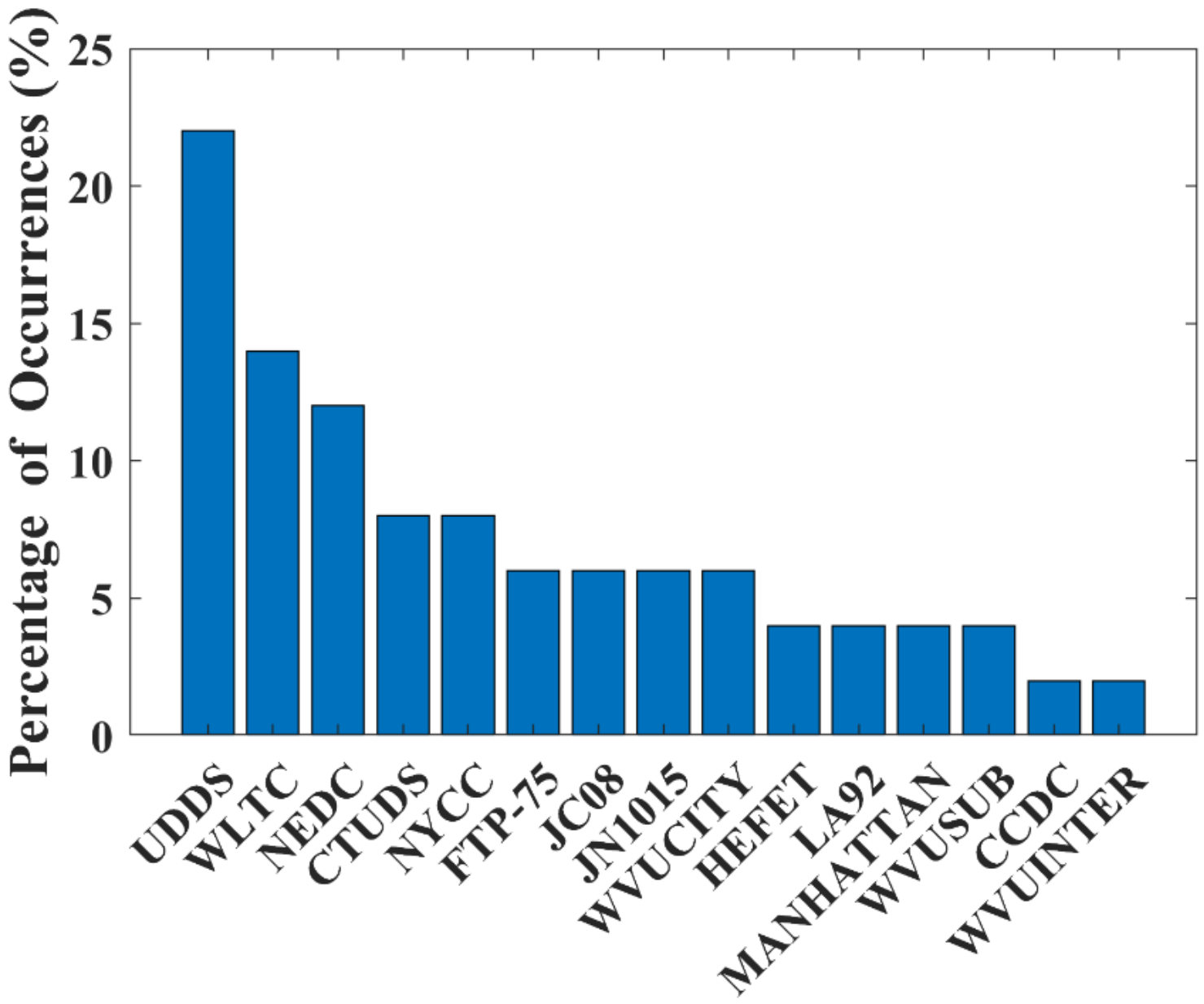

**Fig. 5.** Driving cycles that often occur in RL-EMS studies.

Abbreviation. UDDS: Urban Dynamometer Driving Schedule. WLTC: World Light Vehicle Test Cycle. NEDC: New European Driving Cycle. CTUDC: China Typical Urban Driving Cycle. NYCC: New York City Cycle. FTP-75: Federal Test Procedure. JC08: Japanese JC08. JN1015: Japanese 10-15. WVUCITY: West Virginia University City Driving Cycle. HEFET: Highway Fuel Economy Test. LA92: Los Angeles 92 Driving Cycle. MANHATIAN: Manhattan Driving Cycle. WVUSUB: West Virginia Suburban Driving Cycle. CCDC: China Car Driving Cycle. WVUINTER: West Virginia Interstate Driving Cycle.

## *3.1. Algorithms used in RL-EMS*

From reinforcement algorithms to deep reinforcement learning which has emerged in recent years. Various RL algorithms with their variants are presented with their applications in the MPS-EV EMS field.

### *3.1.1. Early reinforcement learning*

A) Q-learning：

As an early reinforcement learning algorithm, Q-learning is one of the temporal difference (TD) algorithms. As the basis of many RL algorithms, the iteration of the TD method requires only the interaction between agent and environment without modeling and supervision. The learning task is usually to generate optimal control action sequences that lead to maximum rewards [44]. The interaction between the agent and the environment follows a predefined exploration strategy to generate state-action-reward-next state trajectories, typically the ε-greedy strategy. The TD method uses the value-based approach to represent the optimal degree of different state-action under the current policy. The magnitude of the value represents the expectation of subsequent cumulative rewards under the current state-action. To avoid falling into local optimum, the TD algorithms use the Bellman equation to unify the current value with the value of subsequent time steps. The numerical relationship between the current value and the subsequent time-step value is estimated without basing a complex model. In this way, the TD-based algorithms achieve a combination of real-time application and global optimization.

TD can be further divided into single-step temporal difference TD (0) and multi-step temporal difference TD ($\lambda$). In TD (0)-based algorithms, the estimate of this time step requires only the immediate reward and the value of the state-action of the next step, which greatly improves the efficiency:

$$Q(s_t, a_t) = Q(s_t, a_t) + \alpha[r_t + Q(s_{t+1}, a_{t+1}) - Q(s_t, a_t)] \tag{1}$$

where $s_t, a_t, r_t$ denotes the current state, action and reward in the time step t, $Q(s_t, a_t)$ denotes the value of the state-action, $\gamma$ is the discount factor, which reflects the importance of future reward to the current estimate, α denotes the learning rate, which usually decreases as the number of iterations rises.

TD ($\lambda$) needs to obtain reward information and state-action values for the subsequent n time steps to estimate the state-action value of the current time step:

$$\begin{gathered} G_t^{(n)} = \sum_{i=t}^{t+n} \gamma^{i-t} r_{i+1} + Q(s_{t+n}, a_{t+n}) \\ Q(s_t, a_t) = Q(s_t, a_t) + \alpha[(1-\lambda) \sum_{n=1}^{\infty} G_t^{(n)} - Q(s_t, a_t)] \end{gathered} \tag{2}$$

TD (0) makes faster use of data to meet the needs of real-time applications, and TD($\lambda$) iterations are based on more data and are more accurate for value estimation.

The early TD-based algorithms represented by Q-learning stored the estimates of various state-action values in tabular form [45]. Such a table is called Q-table, and the estimate of the state-action stored in the Q-table is called Q-value. The discretization number is limited because of the size limitation of the Q-table and the impossibility of infinite subdivision of action and state variables, which also causes the defect of control accuracy and perceptual accuracy [46].

| | State 1 | State 2 | … | State n |
|---|---|---|---|---|
| Action 1 | -1 | 1 | … | 1 |
| Action 2 | 0 | 2 | … | 0 |
| … | … | … | … | 2 |
| Action n | 1 | 3 | -2 | 0 |

**Fig. 6.** Q-table.

The update of the Q-value is the core of the Q-learning, which is a typical Off-policy algorithm. During the iterative process, the decision of the agent obeys the ε-greedy strategy: the agent has a certain probability of making the action with the highest valuation in the current state or randomly selects the action to explore the impact of the new action on the interaction, and the value of $\varepsilon$ can be adjusted to balance the exploration and exploitation:

$$a = \begin{cases} argmax_a \, Q(s,a) & if \ random > \varepsilon \\ random \ choice & if \ random < \varepsilon \end{cases} \quad (3)$$

However, the equation used to update the Q-value obeys the greedy strategy. The action used for valuation directly takes the action with the highest value in the next state:

$$Q(s_t, a_t) = Q(s_t, a_t) + \alpha[r(s_t, a_t) + \gamma * argmax_{a_{t+1}} Q(s_{t+1}, a_{t+1}) - Q(s_t, a_t)] \quad (4)$$

As a well-established algorithm, Q-learning and its variants are widely used in the research of MPS-EV.

a) Q-learning: Q-learning is one of the TD (0)-based algorithms frequently appearing in RL-EMS research. Xu et al. used Q-learning to develop an EMS for a parallel HEV [47], which improved fuel economy by 8.89% compared to a rule-based approach and 0.88% compared to ECMS. Liu et al. applied the Q-learning algorithm in the energy management of a hybrid electric tracked vehicle (HETV), and the proposed method was compared with the stochastic dynamic planning (SDP) based energy management method. The simulation results showed that the algorithm achieved economic performance beyond SDP with far less training time and a more robust learning capability [48]. Ye et al. used Q-learning to develop an EMS considering battery life and fuel savings to take advantage of supercapacitors handling high power demands, reducing FCEV fuel consumption while slowing down battery degradation [49]. Xu et al. combined multiple Q-learning agents with the same action and reward function settings for forming an ensemble RL framework. Multiple Q-learning agents make action decisions together by weighted averaging. The agents could be integrated during the learning process by sharing rewards. After the train is completed, they could also be integrated to output actions jointly. The results showed that the fuel economy of multiple Q-learning agents was 3.2% higher than the fuel economy of the best single agent [50].

b) Q-learning(λ): Q-learning(λ) is one of the algorithms based on TD(λ), which is one of the first RL algorithms applied to MPS-EV EMS. There was research applying Q-learning(λ) to the energy management of MPS-EV before 2016. Xue et al. used Q-learning(λ) to train the EMS of HEV and PHEV [51]. Two vehicle models were trained under ten working conditions and compared with the rule-based approach. The results showed an average reduction in fuel consumption of 28.8% for HEV and 30.4% for PHEV. The proposed method achieved better economy while the iterations of the algorithm during the training process converged within 3 hours, which is much less than the vehicle service life, demonstrating the high solution efficiency and real-time application potential of the proposed method. The Q-learning(λ) algorithm was also used to manage the energy flow of a hybrid electric energy storage system with a 25% improvement in efficiency compared to a storage system using only lithium batteries [52]. Xue et al. set the battery discharge current and gear ratio as actions to optimize the EMS of parallel HEV and PHEV using Q-learning(λ) and calculate the total driving cost by combining the results of fuel consumption and battery losses. The results showed that the proposed EMS could reduce the operating cost by up to 48% compared to the rule-based EMS [53].

c) Speedy Q-learning: In order to speed up the convergence rate, an algorithm called Fast/Speedy Q-learning has also been introduced into the EMS domain, which modifies the traditional Bellman equation as follows:

$$\begin{aligned} est_1 &= r(s,a) + \gamma max_{a'} Q_{k-1}(s', a') \\ est_2 &= r(s,a) + \gamma max_{a'} Q_k(s', a') \\ Q_{k+1}(s,a) &= Q_k(s,a) + \alpha\big(est_1 - Q_k(s,a)\big) + (1-\alpha)(\, est_2 - est_1) \end{aligned} \quad (5)$$

where $Q_k$、$Q_{k+1}$ denote the valuation of the state-action by Q-tables for the current time step and the next time step, respectively; Du et al. used the proposed method to compare with traditional Q-learning and the experimental results showed that the changes to the Bellman equation improved the convergence efficiency by 16% without affecting the economic performance [54]. Liu et al. also develop EMS based on the Speedy Q-learning algorithm and achieved better training results in comparison with the traditional Q-learning algorithm [55].

B) SARSA：

Like Q-learning, SARSA is a TD-based algorithm that forms policy by estimating state-action values. SARSA also uses the tabular form to store information, and the valuation process is basically the same as Q-learning. The difference is that SARSA is an On-policy algorithm where the agent interacts with the environment using the same strategy as the one used to update the state-action value:

$$Q(s,a) = Q(s,a) + \alpha[r + \gamma * Q(s',a') - Q(s,a)]$$
$$a' = \begin{cases} argmax_a Q(s',a) & if\ random > \varepsilon \\ random\ choice & if\ random < \varepsilon \end{cases} \quad (6)$$

As shown above, in the equation for updating the Q-value, SARSA values the state-action value of the next state and updates the current state-action value according to the ε-greedy strategy.

Chen et al. used the SARSA to solve the MPS-EV EMS problem [56]. Experimental results under real driving cycles showed that the proposed method could significantly improve the fuel economy by up to 21% compared with the conventional charge-depleting, charge-sustaining method, showing great potential in practical applications. The SARSA algorithm was trained to control the fuel cell power of FCEV under different fuel cell switching states, demand torque, and SOC states [57]. Liu et al. also used SARSA for state-action valuation in EMS [58]. Compared with Q-learning, the SARSA valuation is more conservative due to its On-policy feature, which allows SARSA to achieve better training results in some cases. SARSA and Q-learning were used to develop the EMS of an FCEV and were compared with each other [59]. The training results showed that the Sarsa-based EMS kept the battery in better condition for working.

C)Dyna：

The RL agent acquires sequences of states, actions, and rewards through actual or simulated interactions with the environment to learn. An agent learning actual experiences from the environment is called a direct RL agent, and an agent learning simulated experiences from the model is called an indirect RL agent. Direct RL, like Q-learning, learns directly from the experiences generated by the interaction between the agent and the environment. At the same time, Dyna is an artificial intelligence (AI) architecture that integrates learning, planning, and reacting [60]. In Dyna, real experience plays two roles: To refine the agent strategy through learning. It is also used to refine the model to make it more accurate to simulate the environment and subsequently improve the agent strategy based on the environment model.

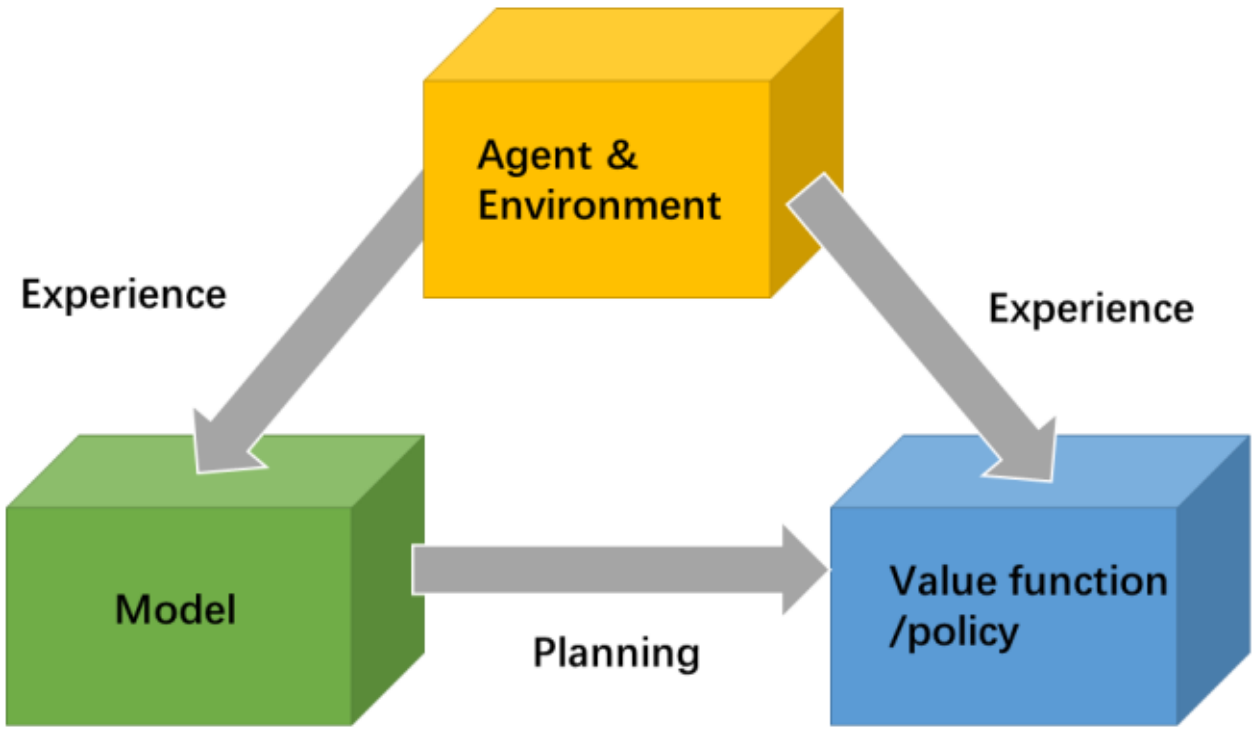

**Fig. 7.** Algorithm Structure of Dyna.

Teng et al. extracted the transition probability matrix of demanded power at different speed ranges based on the experimental driving cycle information and used them to train a Dyna-based EMS [61]. The results showed that the fuel consumption of the EMS trained by Dyna was approximately the same as that of the SDP-based EMS, while the computational cost of Dyna was significantly lower than that of SDP. A related study applied the Dyna-H algorithm by combining the heuristic planning strategy with the Dyna algorithm. Liu et al. applied the proposed method to a parallel HEV, which is closer to the optimization effect of DP than the traditional Q-learning and has a shorter convergence time than Q-learning [62]. Dyna-H was also applied to the EMS development of HETV. The results showed that Dyna-H maintains SOC stability better while achieving a better fuel economy than conventional Dyna [63]. Q-learning, Dyna, and Dyna-H are compared and analyzed in an EMS for PHEV [64], and the authors also compared the training effect of different Dyna planning steps. Yang et al. introduced two new strategies in the Dyna framework, backward focusing and prioritized sweeping, to develop the queue-Dyna algorithm [65]. The Q-learning and Dyna were trained separately to compare with the proposed method. The results showed that the algorithm dramatically improves learning speed and maintains relatively low fuel consumption.

#### *3.1.2. Deep reinforcement learning*

A) Deep Q-network：

Reinforcement learning algorithms such as Q-learning and SARSA discretize the continuous state space and adopt the tabular form to record the estimation of state-action values. However, in complex control problems, where the agent needs to perceive state changes more precisely, the Q-table in a high-dimensional state space would lead to difficulties in storage and search. According to the results of the research on the influence of parameters on Q-learning-based EMS [68], the fuel economy of Q-learning-based EMS decreases when the discretization number of state variables increases, which also proves the shortcomings of Q-learning in terms of perceptual accuracy. Researchers have used neural networks instead of Q-table in control problems to address these problems. The DeepMind team used deep neural networks (DNNs) to fit the state-action value and proposed the deep Q-learning (DQL) algorithm to solve the Atari 2600 game [46]. With this approach, an agent can perceive any slight change in the continuous state space, and the Deep Q-network (DQN) approach is also considered a pioneering work in deep reinforcement learning. With deep neural networks, DQN can handle continuous state spaces. After inputting a state vector, the neural network outputs the state-action values corresponding to different actions under this state, achieving the goal of allowing the agent to control the energy flow in continuous state space. DQN and its variants are widely used in MPS-EV EMS research.

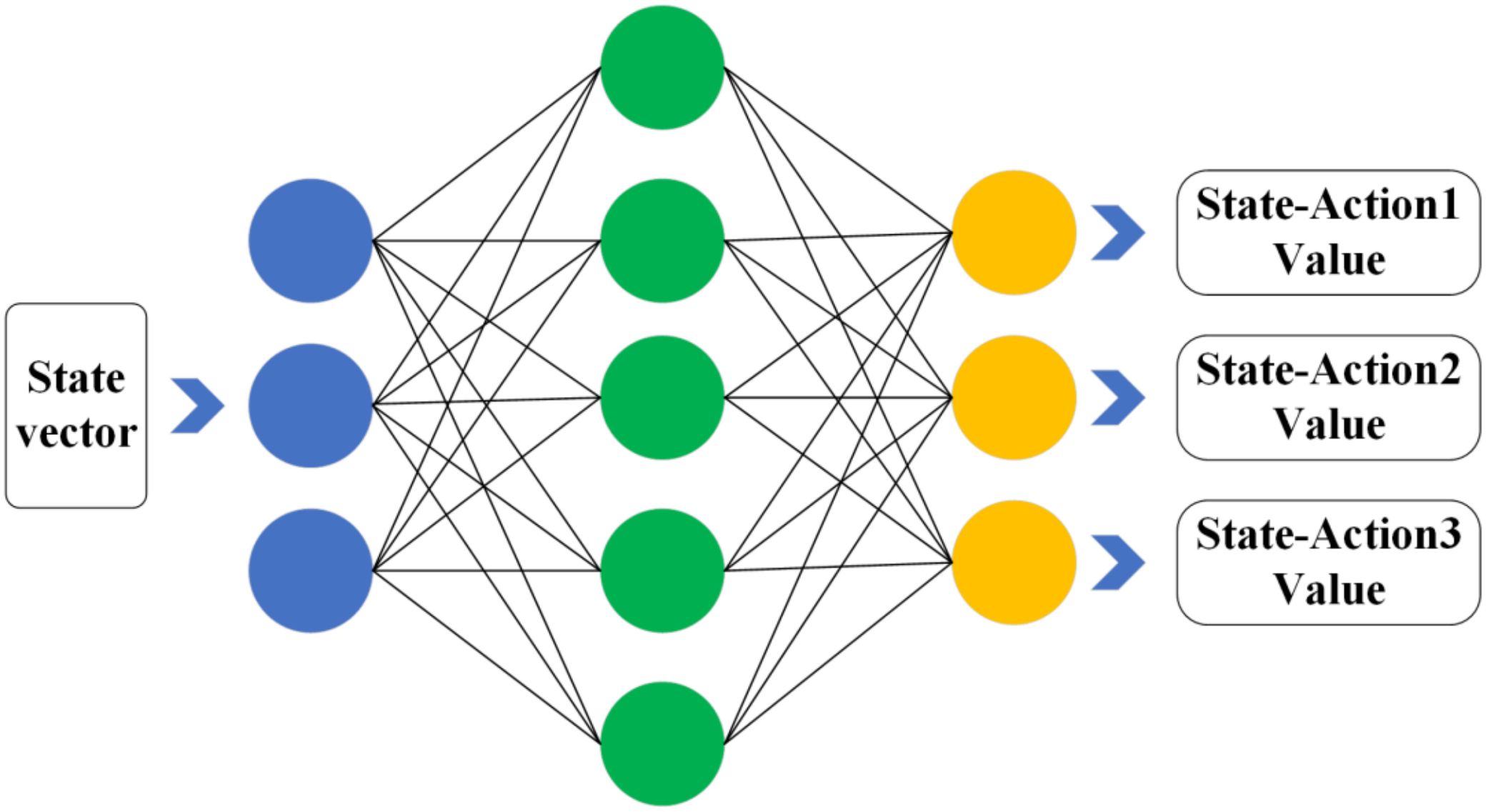

**Fig. 8.** DNN outputs state - action values.

a) The initial version of DQN: The initial version of DQN got the attention of researchers in the EMS field quickly after being invented. Wu et al. used DQN to train EMS [67]. To show the advantages of decision-making in continuous state space, Q-learning based on the same model was also trained for comparison. On the training driving cycle WVUSUB, DQN achieved a fuel economy of 89.6% compared to DP, while the value of Q-learning was 84%. The test conditions are a combination of WVUSUB, JN1015, CTUDC, and other driving cycles. DQN achieves similar fuel economy results to DP while demonstrating the strong adaptability of the DQN algorithm.

b) Double DQN: Double DQN is an improved version that addresses the shortcomings of the initial DQN. During the iterative process, the initial version of the DQN uses the same deep neural network to select the action and evaluate the state-action value for the next step, which leads to a biased estimation of the state-action value. To solve this problem, Double DQN(DDQN) is updated using two DNNs [68], the evaluate network and the target network, respectively. The evaluate network represents the behavioral strategy of the Double DQN, accepts the current state, and makes actions to generate trajectories. These trajectories would be stored in the replay buffer, from which the target and evaluate networks would sample batches for parameter update. The sampling approach of the Double DQN disrupts the order in which trajectories are arranged so that adjacent trajectories are not highly correlated, reducing the variance of updates and improving the efficiency of data use. The target network can be synchronized with the evaluate network by hard or soft updates. The estimates generated by the old parameters can be unaffected by the latest parameters of the evaluate network, thus significantly reducing the divergence and oscillation and improving the stability of the neural network. In view of the above advantages, Double DQN is widely used in MPS-EV EMS.

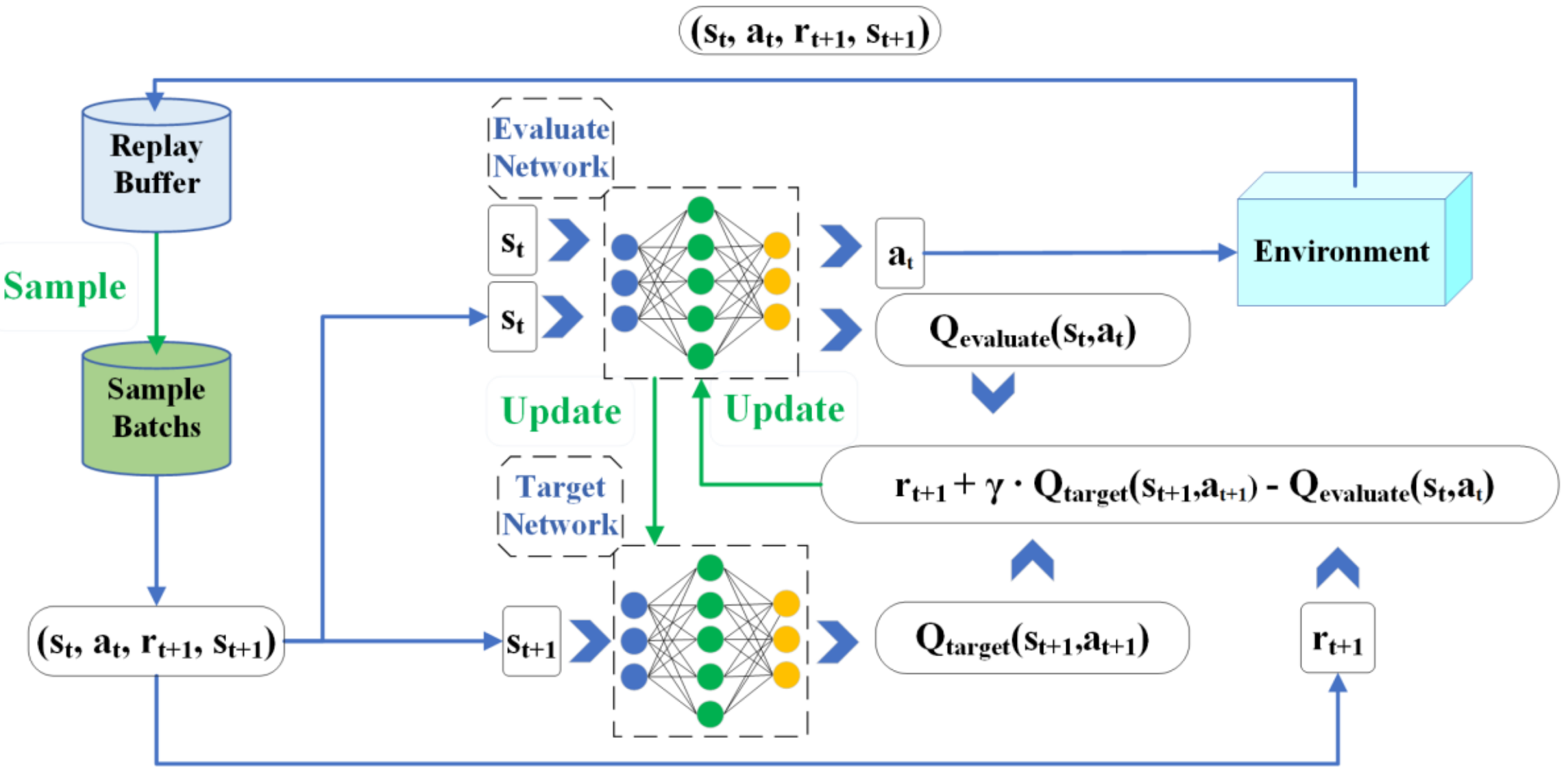

**Fig. 9.** Double DQN algorithm structure.

Han et al. used Double DQN to train a HTEV-based EMS. Compared with the initial version of DQN, Double DQN achieved better results in terms of convergence speed and fuel economy. The validation results on test conditions demonstrate that the improvements made by Double DQN to the initial version of DQN enhance the robustness of the algorithm [69]. The Double DQN-based EMS was compared with the rule-based EMS [70]. The simulation results showed that the Double DQN-based EMS has lower fuel consumption compared to the rule-based EMS, both on training and testing data. Song et al. used Double DQN to develop EMS. The results run on training working conditions showed that the SOC curves of DP and the proposed method follow similar trends, and the difference between them in terms of fuel consumption is less than 6%, showing the ability of Double DQN to approximate the solution of the global optimum [71]. He et al. used Double DQN and Q-learning to verify the improvement of DQL on RL, and the results showed that the DQL-based EMS achieved better economy both in the training and testing driving cycle [72].

While the Double DQN algorithm has achieved good results in simulations, some novel approaches have been introduced into the Double DQN framework to improve the sampling efficiency of the experience and the update efficiency of the network. The heuristic experience replay (HER) is introduced to the framework of Double DQN [73], which achieves more reasonable experience sampling during the iteration of the algorithm. An adaptive moment estimation optimization method with the Nesterov accelerated gradient was also used to update the neural network parameters to accelerate the convergence. Simulation results showed that the proposed method achieved faster training speed and higher fuel economy and approached the global optimum compared to existing DQN methods. A Double DQN with preferred experience playback (PER) was applied to the EMS of FCEV [74], and the learning capability was compared with the standard Double DQN in the research. The results showed that the Double DQN with PER obtained a higher cumulative reward after convergence during the iterative process and kept reward values more stable. Du et al. used a new optimization method, AMSGrad, to update the neural network weights and compared the proposed method with the benchmark method DP and the traditional DQL method [63]. The results showed that the proposed deep reinforcement learning method has a faster training speed and lower fuel consumption than the traditional DQL strategy. Zheng et al. proposed a Double DQN-based EMS method for FCEV [75]. PER was introduced to improve the convergence speed of the DQN algorithm. The proposed method is compared with DP and Q-learning by simulation, and the results showed that the proposed method improves fuel economy by an average of 3.63% compared to Q-learning. The difference in fuel economy between the proposed method and DP is within 6% on average, and the convergence speed of the proposed method is improved by 30.54% on average compared to the Double DQN without PER. The Dueling DQN divides the network output of the original DQN algorithm into two parts: the state-based value and the advantage of the action in this state [76], and calculates the action-independent state values separately, which improves the efficiency of the computation and enhances the robustness of the algorithm. Li et al. used Dueling DQN to develop EMS for HEVs and used priority experience playback to improve fuel economy by 3% compared to standard DQN [77]. Qi. et al. used Dueling DQN and showed that Dueling DQN obtained a faster convergence rate than the Double DQN [78].

B) Deep Deterministic Policy Gradient

Because of the size limitation of the output layer in the neural network, DQN needs to keep the discretization number of actions within a certain range, which also makes DQN unable to handle the continuous control problem and leads to a decrease in control accuracy. Unlike DQN, Deep deterministic policy gradient (DDPG) directly outputs the continuous distribution for the action selection, and the specific value of action would be derived by sampling, and any value of the action may be selected; thus, DDPG solves the problem that deep Q networks cannot be applied to continuous action space [79]. The algorithm flow chart is shown in Fig. 10.

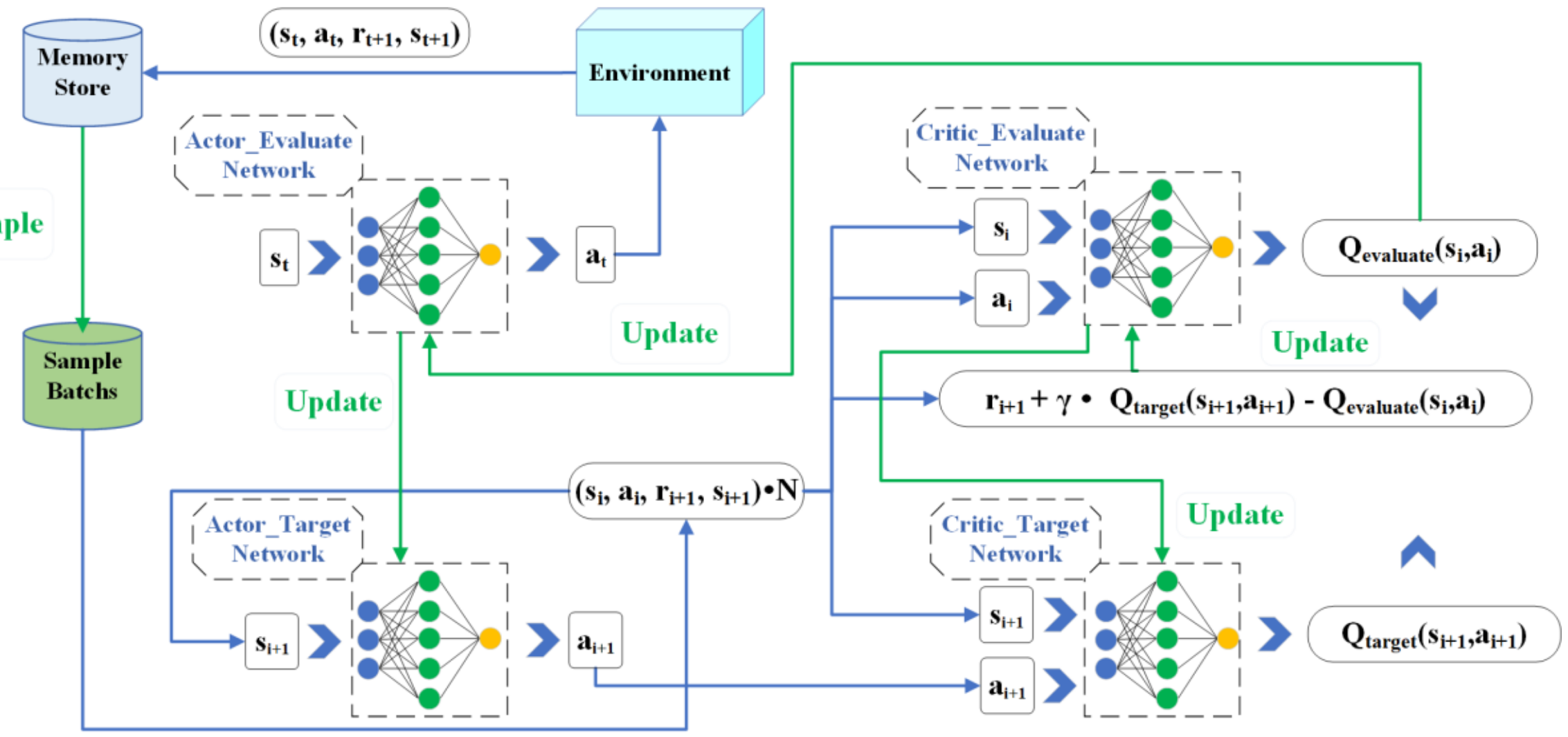

**Fig. 10.** DDPG algorithm structure.

DDPG adopts the actor-critic architecture, where the actor network outputs action, and the critic network estimates the state-action value. Meanwhile, DDPG introduces the target network of DQN with experience replay to reduce the sample correlation and estimation bias.

a) Standard DDPG: With the help of DDPG, related research [36, 37, 80] achieved control over the continuous action space, providing higher control accuracy for energy management problems. The high control accuracy leads to a smooth transmission process in MDP and enhances the robustness of the RL-EMS. Roman et al. used DDPG to develop an EMS, and the training results in all four driving cycles approached the fuel economy of DP, demonstrating that the DDPG algorithm that outputs actions in continuous space can be used to train a better-performing EMS. Ma et al. trained the DDPG network under a real driving cycle. Compared with DP, DDPG achieved 91.3% of fuel economy, while Q-learning only reached 84.9% [81]. Han et al. studied the DDPG algorithm from three aspects: learning ability, generalization, and robustness [82]. The results showed that the fuel economy of DDPG is better than that of double DQN in both test and training data. In addition, the total fuel consumption curve and final SOC curve of each iteration of DDPG during the iteration process have less vibration, indicating that the training effect of the proposed method is more stable than DQN. DDPG was also used to train EMS based on FCEV [83]. During the training process, the neural network parameters converged within 1200 iterations, showing the algorithm's fast learning speed. The fuel consumption under the three driving cycles has reached a gap of less than 10% compared with that of DP.

b) TD3: Twin Delayed Deep Deterministic Policy Gradient (TD3) [84] is an improvement of the deep deterministic policy gradient algorithm. To solve the problem of overestimating state-action values, TD3 uses two critic target networks to estimate the values and chooses the output that is the smaller one of them. The actor network is updated less frequently than the critic network.

TD3 was applied to train EMS for HEV [85]. The hybrid empirical replay (HER) method was introduced based on the mixed experience buffer (MEB). The MEB consists of offline computed optimal experience and online learned experience, which enhances the efficiency of utilizing the learning experience. The improved TD3 EMS obtained the highest fuel economy, the fastest convergence speed, and the highest robustness compared to DQN and DDPG under different driving conditions. In an EMS for HEV [86], the training effectiveness and convergence speed of TD3 and DDPG were compared under two different reward function settings. The results showed that the TD3-based EMS exhibited higher fuel economy and faster convergence speed in the task with a relatively complex reward function. However, in the task with a simple reward function, the fuel economy of the TD3-based EMS was similar to that of the DDPG-based EMS, and TD3 convergence time was much higher than that of DDPG. These results showed that TD3 is more suitable for more complex control problems due to its more complex network structure and process. A TD3-based EMS for FCEV was presented [87]. The TD3-based EMS was trained under randomly generated driving conditions and tested under real driving conditions with different numbers of passengers. The comparison with the benchmark Pontryagin's Minimum Principle (PMP) algorithm showed that the hydrogen consumption of EMS based on TD3 was at most 3.6% higher than that of PMP.

C) Advantage Actor Critic

Advantage Actor Critic (A2C) uses the state value as the benchmark to reduce the variance of the estimate and updates the actor behavior strategy and the critic evaluation function with the state value. The algorithm flow chat is shown in Fig. 11.

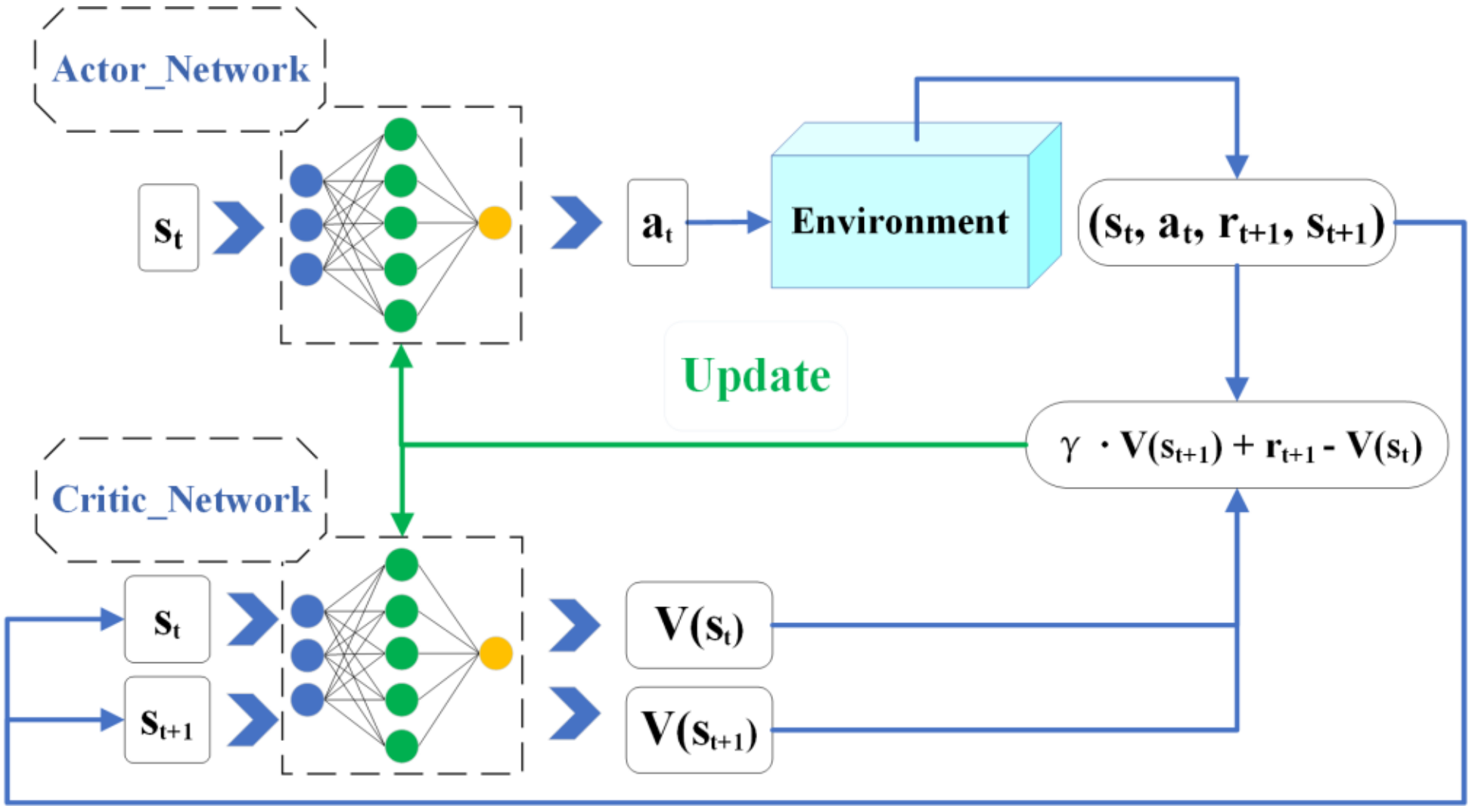

**Fig. 11.** A2C algorithm structure.

Tan et al. applied the A2C algorithm to EMS [88] and compared the proposed algorithm with DP, rule-based, and Q-learning, demonstrating that the A2C algorithm can develop an EMS with high fuel economy and low training cost. In related research, the reinforcement learning framework with a distributed structure was introduced to improve training efficiency in complex problems. Tang et al. proposed EMS with asynchronous advantage actor-critic, compared with standard DQN. Simulation results showed that the distributed A2C algorithm improves the learning efficiency by a factor of 4 compared to DQN [89]. The distributed A2C was used in combination with the Markov chain model (MCM) to develop an online update framework for EMS [90].

D) Proximal Policy Optimization

As an algorithm that can output continuous actions in a continuous state space, the neural network of Proximal Policy Optimization (PPO) [91] outputs continuous actions according to the corresponding probability. Some algorithms perform a completely new sampling after the neural network parameters are updated, causing the problem of low data utilization. To avoid this drawback, PPO sets two actor networks to participate in the update, and the experience obtained by one actor interacting with the environment is used for the update of the other actor. The Kullback-Leibler (KL) divergence is used as the basis to measure the difference between the strategic distribution of the two actors, reusing the accuracy of the learning experience while improving the training speed.

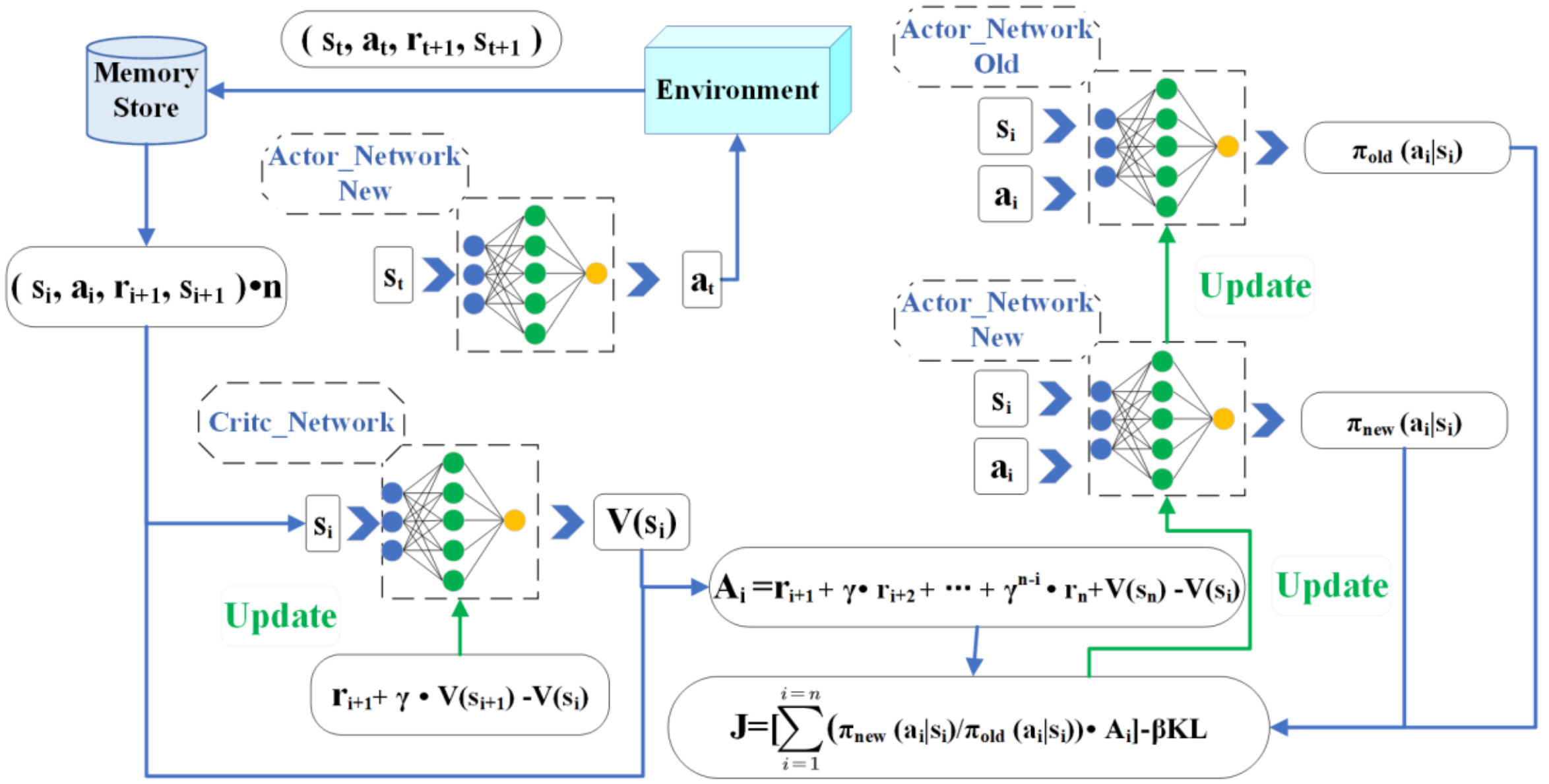

**Fig. 12.** PPO algorithm structure.

EMS based on PPO and TD3 were compared [35]. The results showed that under WLTC, PPO had 9.02% lower fuel consumption than TD3 and lower engine start time and transient operating percentage than TD3. Hofstetter et al. designed a PPO-based EMS for HEV [92]. The average fuel reduction under the tested working conditions was 3.1% compared to the reference strategy, and the pollution emissions were also reduced. Liu et al. used PPO in combination with transfer learning (TL) to develop a real-time EMS [38].

E) Soft Actor Critic

Soft Actor Critic (SAC) is an Off-policy actor-critic deep RL algorithm based on the maximum entropy reinforcement learning framework [93]. The policy optimization objective of SAC requires both increasing the cumulative reward of the agent and making the entropy value of the current policy $\pi_\theta$ as large as possible, which takes the form of:

$$\max_{\pi_\theta} E\left[\sum_t \gamma^t ((r(s_t, a_t) + \alpha\, \mathcal{H}(\pi_\theta(\cdot\,|s_t)))\right] \tag{7}$$

Where $\mathcal{H}(\pi_\theta(\cdot\,|s_t))$ represents the entropy of the current strategy, the larger this value is, the more random the strategy is. $\alpha$ is the entropy regularization factor, which balances the exploration and exploitation of SAC. The significance of the strategy entropy added to the optimization objective is to maximize the reward while making the agent act as flexibly as possible [94].

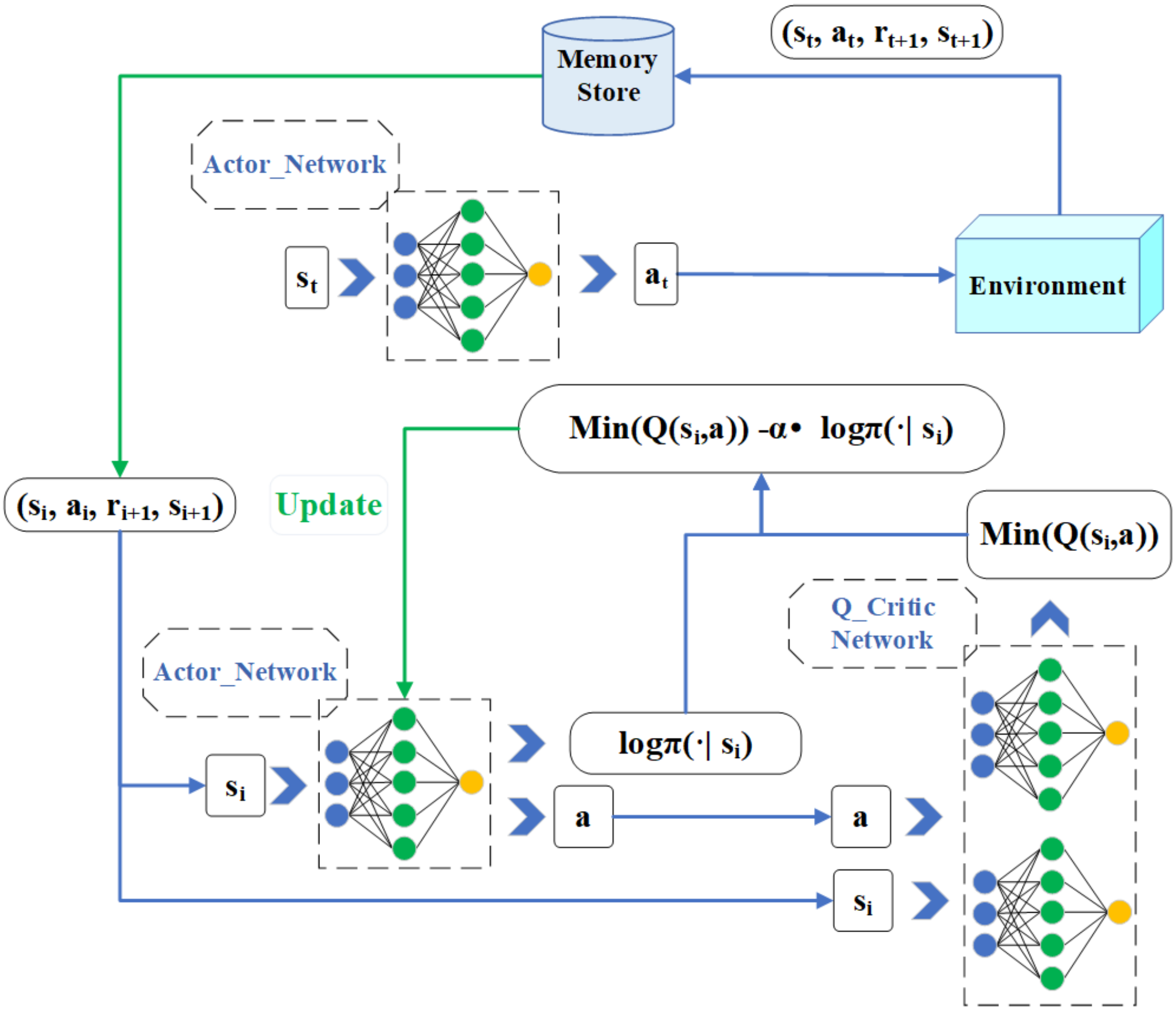

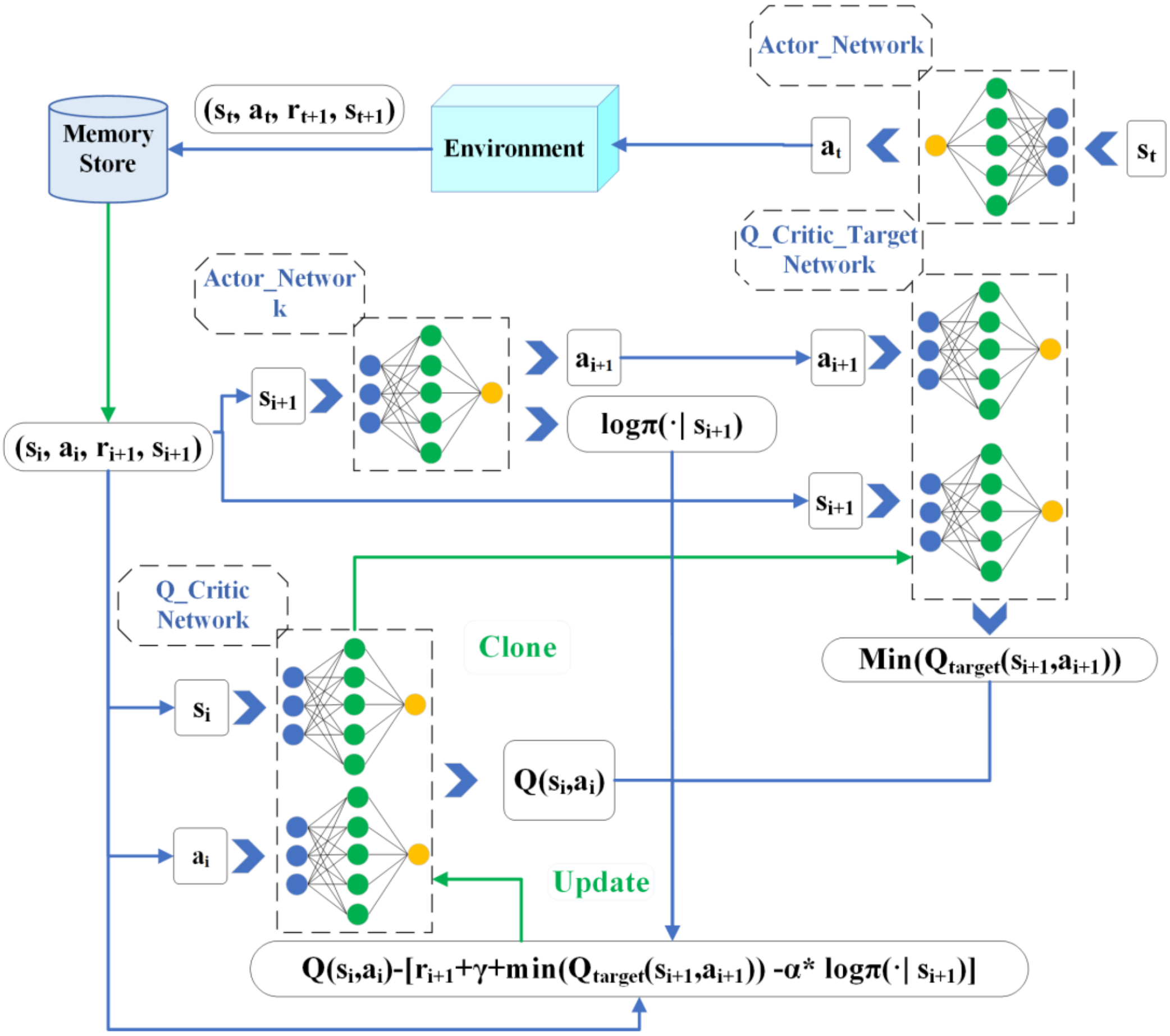

**Fig. 13.** SAC algorithm structure, a) actor, b) critic.

SAC was applied to develop an EMS for hybrid energy storage systems [95] and compared with DQN, DDPG, and DP. In terms of learning capability, the convergence speed of SAC was slightly lower than that of DQN during the iterative process, but SAC had the highest cumulative reward after convergence. The fuel economy performance of SAC under the validation driving cycles was also the best, demonstrating the adaptability of the proposed algorithm. In addition, the computation time of SAC under the validation driving cycles is shorter than the running time of the other two DRL EMSs. These results showed the potential of SAC-based EMS for real-time applications. Wu et al. added the degradation cost of the battery to the available metrics to develop an EMS for HEV using SAC. The research tested the proposed EMS under different driving cycles. The results showed that SAC outperformed the existing DQN method in terms of convergence effect and reduction of overall driving cost [96]. In an EMS for PHEV [97], SAC was compared with DDPG, and the results showed that the SAC-based EMS outperformed the DDPG-based EMS in the economy, and the convergence of SAC was about 36% faster than that of DDPG.

### 3.2. RL-EMS Perception Scheme Design

In MDP, the agent makes actions based on states. The state observed by the agent needs to describe the information needed for decision-making completely. A reasonable perception scheme design can help the agent achieve a more accurate estimation of state-action value and improve the training speed and effect of the algorithm [98]. In the EMS problem, the agent needs to decide on the distribution of energy flow only based on the specific state of the vehicle during the driving process. Deep reinforcement learning algorithms have solved the problem of perceptual accuracy, and the choice of state variables is essential for the perception scheme of advanced reinforcement learning algorithms. The principle of selecting state variables is to represent the noteworthy variable transition during the driving process, enhance the agent's perception, and provide references for control signals. With the development of connectivity technology, Vehicle to Everything (V2X) technology has been utilized to obtain various state variables outside the vehicle to characterize or predict the vehicle's environment. This section explores the state perception schemes prevalent in RL-EMS research in terms of both in-vehicle and out-of-vehicle information.

Table 2. Vehicle state perception scheme.

| Information sources | | State variable | Reference |
|---|---|---|---|
| In-vehicle information | | Speed | [50, 51, 53, 56] |
| | | Acceleration | [37, 80, 82,87] |
| | | Power demand | [51, 56, 63, 65] |
| | | Torque demand | [53, 50, 71, 36] |
| | | SOC | [56, 58, 70, 77] |
| | | Gear ratio | [50, 66, 36] |
| | | Speed of generator | [48, 54, 55] |
| Out-of-vehicle information | V2V | Distance from surrounding vehicles | [102, 103, 104] |
| | | Speed/acceleration of surrounding vehicles | [102, 103, 104] |
| | V2I | Traffic light | [106, 103, 104] |
| | V2N | GPS location | [99, 108, 12] |
| | | Terrain | [109] |

#### 3.2.1. In-vehicle information

The relevant physical quantities of the vehicle itself are reflections of the vehicle's working condition, as shown in Table 2. The most intuitive and commonly used state variables inside the vehicle are speed [50, 51, 53, 56], acceleration [37, 80, 82, 87], SOC [50, 66, 67, 77] and demand power [51, 56, 63, 65]. Since these variables are related to each other, the RL-EMS can perceive the working condition of the vehicle by selecting only a few of them. The choice of state variables varies depending on the vehicle type. For example, an EMS for a Plug-in Hybrid Electric Bus (PHEB) [99] chose to record the number of passengers in the vehicle as a state. Some special vehicles require different state variable settings. For example, in an EMS for the HETV, the state variables include SOC, demand power, longitudinal speed, and angular speed [69]. On the premise that the state of the powertrain is completely described, the settings of state variables inside the vehicle usually do not differ significantly between vehicles of the same type.

#### 3.2.2. Out-of-vehicle information

With the development of vehicular networking technology and related research, as shown in Fig. 14, vehicle perception of the road environment provided by Vehicle-to-everything (V2X) information began to guide MPS-EV energy management, and a series of information outside the vehicle has been introduced into the state variables. For example, Chen et al. collected a large number of high-definition images of five typical roads in the racing game Dust Rally 2.0 to build an environment model containing a series of variables such as driving images, road gradient, longitudinal speed, and the number of passengers [100]. The introduction of the ability to perceive information outside the vehicle has driven research on the effect of outside information on EMS.

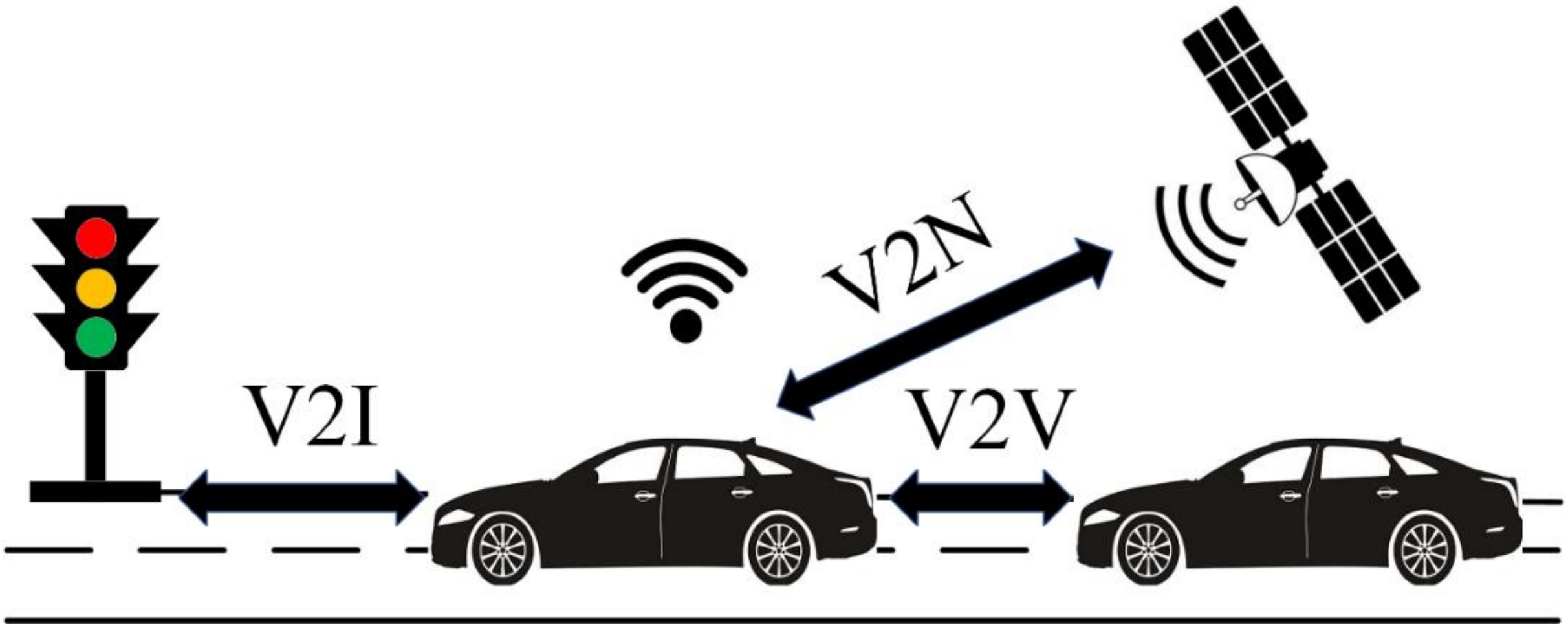

**Fig. 14.** V2X information.

A) V2V information

Vehicle to Vehicle (V2V) technology is not limited to fixed base stations and provides direct end-to-end wireless communication for vehicles on the move. That is, with V2V communication technology, vehicle terminals exchange wireless information directly with each other without forwarding through a base station [101]. The use of V2V information by RL-EMS focuses on obtaining the distance between vehicles and the speed of other vehicles in the traffic flow, both of which can influence the estimation of the vehicle's future speed, showing the current and expected speed magnitude of the MDP.

Li et al. incorporated preceding vehicle speed, preceding vehicle acceleration, and distance to the preceding vehicle into the state variables of a DDPG-based EMS [102]. Inuzuka et al. proposed an EMS introducing road congestion into the perception scheme. Distance to the preceding vehicle, preceding vehicle speed, preceding vehicle acceleration, distance between the vehicle in front and two vehicles ahead, speed of two vehicles ahead, and acceleration of two vehicles ahead were obtained as state variables. The traffic congestion level was quantified as a state for the reference of the agent decision-making [103]. He et al. proposed a cyber-physical system (CPS)-based EMS, calculated the average distance between vehicles in the traffic flow and the speed of the vehicle ahead by V2V information, which is used as a state variable to represent the driving condition. The experimental results showed that the effectiveness of both DDPG and DQN was improved after the introduction of traffic information [104].

B) V2I information

Vehicle to Infrastructure (V2I) is the exchange of information between vehicles and road infrastructure. V2I technology allows vehicles to communicate with some road infrastructure in order to transmit the speed and location of the vehicle to the central server. The server records the speed and location of all vehicles and provides this data for some applications [105]. With V2I information, vehicles can be informed of traffic signals and the traffic conditions of the road ahead.

In RL-EMS, V2I technology is mainly used to obtain the signals of traffic lights on the road and the distance of vehicles from the next traffic light. The macroscopic traffic sensing capability of RL-EMS is enhanced by using V2I information as the state variables. In an EMS for HEV [103], traffic information was input to the deep neural network as a basis for decision-making. The state variables include traffic light signals, number of seconds, and distance of the vehicle from the next traffic light. The comparison showed that the introduction of traffic information improves the calculation speed, and the fuel economy of the proposed method is better than that of MPC. V2I information such as traffic light signals and their timing, distance from the traffic light were obtained as state variables for an EMS [104]. The onboard computer vision hardware and software used the detected visual information, such as traffic lights, as the DDPG network's state input [106]. The EMS fuel consumption with visual information was reduced by 4.3 ~ 8.8% compared to the DDPG without visual information.

C) V2N information

Vehicle to Network (V2N) refers to the connection of in-vehicle devices to a cloud platform through an access network. The cloud platform and the vehicle interact with each other, and the cloud platform stores and processes the acquired data to provide various application services needed by the vehicle [107]. In the RL-EMS study, V2N technology is mainly applied to obtain the vehicle position in Global Positioning System (GPS).

For a given trip, future energy consumption is highly correlated with the remaining distance. Some researchers have used GPS to obtain the vehicle's location during the trip to guide energy management. As the PHEB with the most stable working conditions, the position on the line it is on can reflect the characteristics of the environment at this time to the agent to a great extent. Therefore, in an EMS for PHEB [99], the current line of the vehicle was added to the state variables. The proposed method achieved an effect close to DP in fuel economy and showed good generalization performance in the test training driving cycles. Liu et al. proposed a reinforcement learning method incorporating trip information for the characteristics of PHEV. The algorithm obtained the remaining travel distance from GPS for energy optimization [108]. The introduction of the remaining travel distance provided a more effective basis for the agent to control the SOC. The proposed method achieved better training effects than the rule-based method while maintaining the SOC level better. Li et al. integrated historical cumulative trip information into a DDPG-based EMS for practical SOC guidance. The proposed method is systematically introduced from offline training to the online application, and the DP method is used to obtain the SOC trends of working conditions that match the vehicle driving habits. The travel distance after the last charge is added as a state variable to guide the SOC. Compared with MPC-based EMS, the proposed EMS is more adaptive and robust to dynamic driving conditions, and the proposed method could also learn to predict future travel information from historical data [12]. Due to the high mileage and load of commercial vehicles, terrain information was incorporated into the state variables of hybrid commercial vehicles to train the DDPG-based EMS [109]. Experimental results showed that the fuel gap between the proposed method and the DP-based method is around 6%. The inclusion of terrain information improves the fuel economy of the proposed method by approximately 2%.

### *3.3. RL-EMS Decision Scheme Design*

The decision scheme of RL-EMS influences the state transition of MDP. There are different control frameworks of RL algorithms to directly determine or indirectly influence the energy distribution of the MPS-EV. The control frameworks studied in RL-EMS are summarized into three: end-to-end control, hierarchical control, and parallel control.

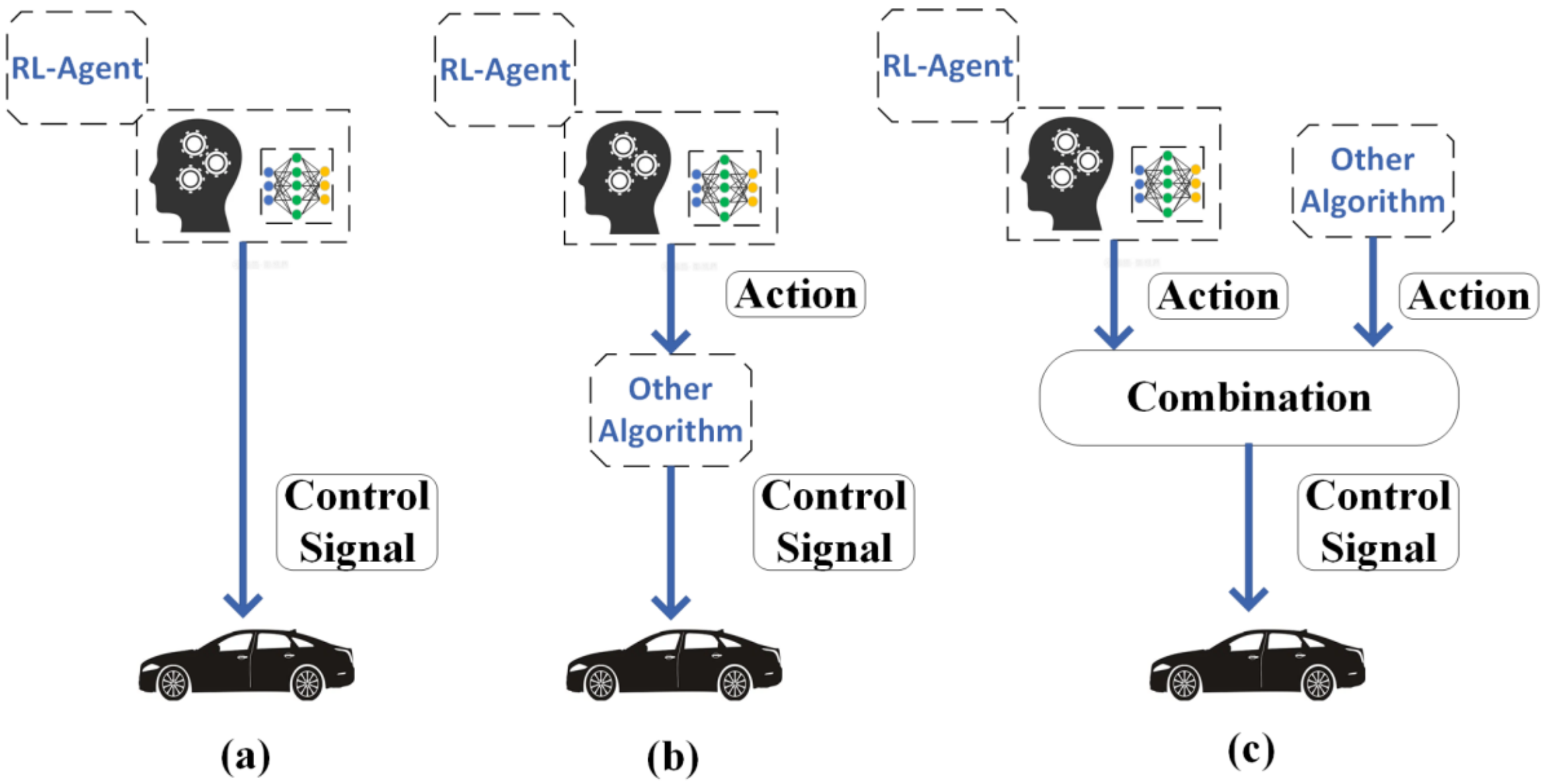

**Fig. 15.** Control frameworks in RL-EMS.

### *3.3.1. End-to-end framework*

Most of the RL-EMS research mentioned in this paper uses an end-to-end control framework. For the RL-EMS decision scheme using the end-to-end control framework, the actions made by the agent can be directly used as the final control signal to decide the energy distribution of the vehicle. In the energy management problem solved by the end-to-end control framework, the control of energy flow by RL-EMS can be divided into continuous and discrete action control. The agent can only pick its action from several predefined values for discrete action control. The control accuracy of the EMS is directly related to the discretization number [66]. The agent can output any value taken in the range of continuous action space for continuous control. The deep reinforcement learning algorithm for continuous control is achieved by sampling over the normal distribution, where the DNN outputs the mean and variance of the normal distribution, which also determines the action distribution of the agent. Then the final action is then determined through sampling [80].

A) Discrete control solution

In RL-EMS research that adopts discrete control, different kinds of physical quantities are discretized. Engine torque [70, 71], motor torque [47, 50, 66], and engine power [65, 78] are the action variables commonly used. In an EMS for HEV [66], the motor torque was chosen as the action and was limited to the range of -100 to 250. To verify the effect of control accuracy improvement on the training effectiveness of the Q-learning algorithm, the motor torque was uniformly discretized into different numbers, 5, 10, 20, 50, and 100. Experimental results showed that increasing the discretization number enhanced the training effectiveness. However, as the discretization number increases, the training effect enhancement becomes smaller. In another Q-learning-based EMS [47], the motor torque was uniformly discretized into 100 values from -90 to +90. In DQN, the limitation of the output layers in the neural network does not allow a too larger number of discretization. Engine torque is selected as action in a DQN-based EMS with a discretization number of 24 [70]. In another DQN-based EMS [71], the action size of the DQN neural network is 10, and the engine torque is uniformly discrete from 0 to 207. Engine power is also often selected as the action in the MPS-EV EMS. Qi et al. defined the actions as 24 power level output from the engine [78]. In an FCEV-based EMS [59], the degree of hybridization (DOH), which is the ratio of fuel cell power to demand power, was uniformly discretized from 0.1 to 0.9 as ten values for the actions of the SARSA algorithm.

The increase in discretization number helps to cover different intervals of the action space and to improve the control accuracy. An alternative approach was used to refine the energy distribution of the vehicle while meeting the dimensional requirements. The deep neural network outputs the action not as a specific value but as the change in the current physical quantity, and the amount of change in the engine power is set to six options [67]. This approach avoids excessive deep neural network dimensionality while achieving high control accuracy. Han et al. used the amount of change in engine speed as the action, selected from eight options [69]. On FCEV-based discrete control, the amount of change in fuel cell power was also used as the action [74].

B) Continuous control solution

The motor torque was selected as the action on one EMS [36]. With the help of the DDPG algorithm, the control of continuous action space is implemented. Guo et al. selected the engine power as action [80]. Ma et al. selected the throttle opening of the engine as action [81]. Tang et al. selected the amount of change in engine power as the control action, which ranges from -5 to +5 [89]. The smaller range avoids too fast variation of engine power and ensures that the engine operates in a stable working condition. Most of the FCEV-based reinforcement learning algorithms choose fuel cell power as action [87, 34].

### *3.3.2. Hierarchical control and parallel control in RL-EMS*

A) Hierarchical control

As shown in Fig. 15, in RL-EMS with a hierarchical control framework, the RL agent does not directly output actions that directly determine energy distribution, such as engine torque. The upper layer of the framework outputs relevant actions as the reference of the lower layer, and the final control signals are output by the lower control strategy.

Li et al. constructed a hierarchical control framework by DDPG and A-ECMS. In the upper layer, the reinforcement learning agent planned the reference vehicle speed in real-time based on the traffic information and powertrain state. Then the A-ECMS at the lower layer could optimize the energy distribution based on the demand torque determined by the reference vehicle speed [102]. In an HEV-based EMS [110], the ECMS and RL were combined to develop a hierarchical control framework. The agent outputs a new co-state, which is similar to the equivalence factor, based on SOC, current co-state, torque demand, predicted power demand, predicted average vehicle speed, and predicted average acceleration. The lower layer decides the energy distribution of the vehicle based on the co-state. The experimental results showed that the improvement of the proposed method outperformed A-ECMS in terms of economy. A new hierarchical control framework was applied to the energy management problem of FCEV [57], in which both the upper and lower layers used reinforcement learning algorithms to output actions. The upper layer output the penalty function through the segment average vehicle speed, the trip average vehicle speed, the initial SOC, and the segment distance. The penalty function affected the optimization objective of the lower layer strategy. The lower layer strategy output the control signal to directly decide the fuel cell power through the penalty function from the upper layer and state variables such as SOC and demand power. Zhou et al. developed an EMS for FCEV based on reinforcement learning and thermostatic control. The RL strategy output upper and lower limits of the battery SOC, and thermostatic control controlled the energy distribution based on the SOC range determined by reinforcement learning. The results showed that the combination of reinforcement learning and thermostatic temperature control ensured that the fuel cell could achieve high fuel economy with stable output power [111]. Chen et al. trained an effective and convergent reinforcement learning controller based on the demand power distribution under multiple working conditions. The reinforcement controller was embedded into a stochastic MPC controller by constructing a multi-step Markov speed prediction model to determine the optimal power of the battery in the prediction time domain. A control flow similar to the hierarchical control framework was proposed [112]. The optimal brake specific fuel consumption curve of the engine was considered the prior knowledge as the upper layer strategy, while the RL algorithm was used for the lower layer. The upper layer strategy determined the appropriate action space for the RL agent in the current state, which simplified the action space of the engine and increased its average efficiency. The RL agent did not need to explore along the global map of the engine, which allowed the algorithm to search for the optimal solution in a smaller space. Compared with conventional RL, the proposed method achieved a better training effect with less computational cost.

B) Parallel control

As shown in Fig. 15, in the parallel control framework, the control signals are derived from the collaborative decision-making of different strategies. RL agent outputs actions that provide control signals in parallel with other controllers in the control framework. Wu et al. combined MPC and RL algorithms to develop an EMS [113], with each algorithm outputting actions to control the energy distribution. The fuzzy logic controller output adjustment factors to determine the linear weights of MPC and RL actions to determine the final control signal. Tang et al. proposed a parallel framework for providing control signals by two reinforcement learning algorithms [114]. The DQN agent learned the gearshift strategy to determine the transmission ratio and the DDPG agent controlled the throttle opening of the engine.

### *3.4. Reward functions in RL-EMS*

RL-EMS aims to form a policy that can output a sequence of actions for the working condition. The ultimate goal of the action sequence is to maximize the reward accumulated for the driving cycle. The setting of the reward function establishes the policy goal of the agent in the MDP [21]. For the MPS-EV EMS problem, the reward function setting needs to achieve at least the goals of reducing fuel consumption and maintaining SOC. In order to quantify the driving cost and improve economic performance, different metrics are applied to the reward function. Different reward function settings are categorized and presented as follows.

#### *3.4.1. Fuel consumption*

To avoid specifying weights for different types of energy consumption, Song et al. optimized only fuel consumption. The environment rewarded the agent negatively when the power exceeded the physical limit of SOC to balance the energy distribution among energy sources [71].

$$\begin{cases} r = -fuel & if\ \left|SOC - SOC_{final}\right| < 0.05 \\ r = -SOC_{penalty} & if\ \left|SOC - SOC_{final}\right| > 0.05 \end{cases} \tag{8}$$

Where $SOC_{penalty}$ represents the high negative reward and $SOC_{final}$ represents the final value of SOC pre-determined by the researcher. Yue et al. set up the reward function in this form [70]:

$$r = \begin{cases} \dfrac{1}{fuel} & fuel \neq 0 \cap 0.4 \leq\ SOC \leq 0.85 \\ \dfrac{1}{fuel + C} & fuel \neq 0 \cap\ SOC < 0.4\ or\ SOC > 0.8 \\ \dfrac{1}{Min_{fuel}} & fuel = 0 \cap 0.4 \leq\ SOC \\ -\dfrac{1}{C} & fuel = 0 \cap\ SOC < 0.4 \end{cases} \tag{9}$$

Where $C$ is the numerical penalty when the SOC exceeds the limit and $Min_{fuel}$ is the minimum non-zero value of the instantaneous fuel consumption value. The common feature of these reward functions is that they only quantify the instantaneous fuel consumption, but when the SOC cannot be maintained, a larger negative reward will be fed back to the agent.

### *3.4.2. Equivalent fuel consumption*

Related research [47, 66, 56] drew on the idea of ECMS to assign weights to different types of energy consumption. The instantaneous equivalent fuel consumption is incorporated into the reward function as follows:

$$r = a - (\gamma * fuel + \delta * elec) \tag{10}$$

Where $a$ is a constant, $fuel$ is the fuel consumption in one time step, $elec$ is the electricity consumption in one time step, and $\gamma$， $\delta$ are the corresponding weights. Such a setup also faces the problem inherent to the ECMS algorithm that the weights need to be set artificially beforehand. When the values do not correspond to the current driving cycle, the training effectiveness of the algorithm will be reduced.

### *3.4.3. Variation of SOC*

In the RL-EMS research mentioned in this paper, the combination of fuel consumption and SOC variation is the most widely used design in the reward function. The common reward function settings are as follows [48, 54, 55, 58, 61, 62, 65, 77]:

$$r = b - (\gamma * fuel + \beta * \Delta SOC^n) \tag{11}$$

Where $b$, $n$ are constants, the maintenance of SOC can be directly reflected in the reward function, $\Delta SOC$ represents the difference between the battery power and the reference SOC, and $\beta$ is the corresponding weight. However, the weights still need to be specified in advance, and the training effect is highly dependent on their values. Lian et al. investigated the effect of different weights on the training effect in a DDPG-based EMS [112]. γ was fixed at 1, and 8 values in the range of 105 to 3500 were selected and assigned to $\beta$. The results showed that values within a reasonable range can improve the fuel economy to more than 93% of the fuel economy of DP and that too high and too low values cannot achieve such control effect. Values that are too low do not maintain power well, and values that are too high waste the buffering effect of the battery.

Kai et al. proposed a TD3-based energy management strategy for FCEV considering fuel cell aging and adopted a new reward function to maintain power and protect the battery [34]. The training results showed that during the training process, the new reward function setting could accelerate the convergence, and the total reward after convergence was higher:

$$G_{bat} = \omega_{bat} * \Delta SOC * I_{bat} \tag{12}$$

where $G_{bat}$ represents the power consumption part of the reward function. $I_{bat}$ represents the battery current, and $\omega_{bat}$ is the corresponding weight. Such a setting not only reflects the current SOC, but also takes the instantaneous energy consumption of the battery into consideration. Han et al. introduced a new reward function by combining the equivalent fuel consumption with SOC [82].

$$r = -[\alpha(fuel + elec) + \beta * \Delta SOC^4] \tag{13}$$

### *3.4.4. Health of components*

In addition to the energy consumption of different sources, component health can also be seen as an influence on driving costs, so component maintenance also appears in some reward functions of RL-EMS, especially for FCEV-based EMS. As the core component of the FCEV powertrain, the fuel cell service life is strongly influenced by the energy distribution strategy, and excessive output power variation is detrimental to its health.

Xt et al. constructed a model to quantify the degradation of the fuel cell and introduced it into the reward function [74]:

$$r = -[\alpha * fuel + \beta * (SOC - 0.7)^2 + \omega D_{FC,deg}] \tag{14}$$

Where $D_{FC,deg}$, $\omega$ represents the degree of fuel cell degradation and the corresponding weights. Li et al. incorporated fluctuations in fuel cell output power into the reward function [143]:

$$r = \begin{cases} \frac{1}{m_H + \omega|\Delta P_{fc}|} & m_H \neq 0 \cap SOC_{min} \leq SOC \leq SOC_{max} \\ \frac{1}{m_H + 10 + \omega|\Delta P_{fc}|} & m_H \neq 0 \cap SOC_{min} < 0.4 \ or \ SOC > SOC_{max} \\ \frac{1}{SOC + 1} & m_H = 0 \cap SOC \geq SOC_{min} \\ \frac{1}{P_{batt}} & m_H = 0 \cap SOC < SOC_{min} \end{cases} \tag{15}$$

Where $m_H$ represents the fuel consumption, $P_{fc}$, $P_{batt}$ represents the fuel cell, and battery power. When the fuel cell power varies too much, the reward value decreases accordingly. In related studies, the temperature of the battery is also limited to prevent the over-temperature [96]:

$$r_{tem} = \begin{cases} 0 & if \ T < T_{tar} \\ \tau(T - T_{tar})^2 & if \ T \geq T_{tar} \end{cases} \tag{16}$$

Where $r_{tem}$ represents the part of the reward function related to the battery temperature, T represents the battery temperature, $T_{tar}$ represents the reference battery temperature, and $\tau$ represents the corresponding weight. In addition to the fuel cell, the engine service life is also strongly influenced by energy distribution. The reward function is subtracted by a fixed constant when frequent start occurs [112].

### *3.5 Innovative training methods of RL -EMS*

After determining the algorithm to be used and completing the modeling of the main factors of the MDP, the RL algorithm can be applied to the EMS problem of MPS-EV. However, unlike some control problems, in the energy management of MPS-EV, the working conditions of the controlled vehicle are not immutable, and the agent's environment has different styles, which affects the training effectiveness of the proposed method. The long exploration time of the agent to the environment also reduces the training efficiency. In order to make the reinforcement learning closer to the actual conditions of the vehicle, when building the RL-based EMS, some researchers adopted innovative training methods to enhance the adaptability of agents to the changing environment, and some other researchers shortened the training time of the agent by effectively initializing the behavioral strategies.

Table 3 Innovative training methods in RL-EMS.

| Goal | Way | Method | Reference |
|---|---|---|---|
| Adaptability to changing environment | Diversity of training data / driving cycles | Generate training data online | [36, 123] |
| | | Randomly add other driving cycles | [117] |
| | Characterize driving cycle characteristics to output corresponding strategies | Mark trip distance | [119, 118] |
| | | Mark speed | [57,80] |
| | Update online when driving style changes | Introducing state transition probability matrix | [55, 120, 121, 122, 123] |
| Initialization of agent policy | Initializing the Q-table by expertise | Give higher value to high-efficiency actions | [126] |
| | | Remove low-efficiency state-actions | [127] |
| | Initializing the network based on TL | Get parameters from other EMS | [37,38] |

#### *3.5.1 Adaptability to the changing environment*

The nature of the reinforcement learning algorithm, which relies on a large number of samples, dictates the need to train the agent in a simulation environment and then migrate it to the actual application. In order to be successfully applied in the real world, the RL-EMS needs to have the ability to handle various working conditions to ensure that the strategy can achieve good economics even when the data distribution of the training data is somewhat different from that of the testing data. Xu et al. explored the adaptability of an HEV-based EMS developed through Q-learning [115]. Compared with ECMS, MPC, and thermostatic control strategies, experimental results proved the self-learning capability of Q-learning when driving conditions, vehicle loading conditions, and road classes change. Q-learning could adapt to the adjustment of working conditions and change the strategy as fast as possible during iterations, demonstrating the agent's adaptability to changing driving conditions due to the self-learning capability of RL. Related research [72, 116] applied EMS based on training data to another style of testing data to verify the adaptability of the RL algorithm applied to EMS. The results showed that the trained method adapted well to driving cycles with different styles. Inuzuka et al. prepared testing data with similar characteristics to the training data and testing data with distinct characteristics to the training data in the simulation validation [103]. The results showed that the testing effect was better when the testing driving cycle was similar to the training driving cycle. This result led to the conclusion that increasing the diversity of training data styles helps improve the proposed method's generality. Zhu et al. developed a model with mixed urban road conditions and traffic density in the Simulation of Urban Mobility to automatically generate training data that provides sufficient variability in the vehicle speed profile [117]. Liessner et al. added the unique random cycle to the driving cycle alternating training [36]. The results showed that the addition of noise barely affected fuel economy performance on the training data but achieved much-improved fuel economy performance on the testing data.

Different styles of driving cycles have different state transition processes. The diversity of training data can undoubtedly enhance the adaptability of RL-EMS. However, during the application process, the agent also needs to identify the current driving style in order to achieve the goal of accelerating the policy update. When the driving style changes, the agent needs to make adjustments to accelerate learning the new driving style. Wang et al. enhanced the diversity of training driving cycles to accommodate trips with different distances and average speeds and characterized trip characteristics in detail. A series of information such as preset driving distance, driven distance, and driven time are used to guide the EMS in making decisions. In the test driving cycles, the proposed approach achieved an average fuel efficiency improvement of 21.8% compared to the rule-based method [118]. In a PHEV-based EMS, the RL algorithm combines neural dynamic planning (NDP) with future travel information to estimate the expected travel cost for the given vehicle. The driving cycle is characterized in terms of travel distance. Two machine learning algorithms, Q-Learning-Short Trip (QL-ST) and Q-Learning-Long Trip (QL-LT) were developed for energy control of the PHEV for short-distance and long-distance working conditions, respectively [119]. Guo et al. used transfer learning (TL) to adapt to different speed intervals of the controlled vehicle [80]. When the speed distribution of the training driving cycle differs significantly from that of the actual driving cycle, it may take a long time to build a new EMS. To solve this problem, a bi-level control framework was constructed. Since speed can represent driving style to a large extent, the algorithm can adopt similar behavioral strategies on different driving cycles with little difference in speed. The upper layer used the DDPG algorithm for EMS training on different driving cycles to develop three DDPG deep neural networks based on the division of speed intervals. The TL approach was then used to initialize the neural network for a new driving cycle, and knowledge transfer was achieved between similar speed intervals of different driving cycles. The results showed that the TL-initialized DDPG algorithm saves about 0.1 SOC and 9.9% fuel compared with the standard DDPG algorithm on the testing driving cycles, which verifies the effectiveness and adaptability of the framework. Yuan et al. classified frequently occurring routes based on the historical working conditions of the vehicle into segments by road location and traffic condition, obtained the long-term average speed of vehicles by k-nearest neighbor method and predicted short-term speed sequences by the new model averaging method. The characteristics of the current working conditions are extracted to enhance the ability to perceive and adapt to the working conditions [57].

Related research [55,120,121] chose to characterize the current working conditions in terms of state transition probability matrix rather than speed ranges. In such research, state transition probability matrices were generated for each time period. The induced matrix norm was used to quantify the difference between working conditions, and the corresponding strategies were adjusted when the driving style was judged to have changed. The proposed energy management strategy significantly improves fuel economy and convergence speed compared to conventional RL. Lin et al. used Markov chains to calculate the transfer probability matrix of demand power with KL divergence to measure the difference between current and past working conditions, which allows the EMS to be updated with driving cycles. The computational cost and economic performance of RL-EMS with different KL divergence update thresholds were analyzed. The proposed method was compared with DP and rule-based and showed a great training effect [122]. An online update strategy is proposed [123]. RL and MPC are combined to design an online update strategy for a line-fixed HEV. The study generates the state transfer probability matrix based on Markov chains to predict future working conditions and adjust them in time. The algorithm uses experience replay to accelerate the update of network parameters when the driving condition style changes.

### *3.5.2 Initialization of agent policy*

RL-EMS, as a data-driven unsupervised AI method, requires a long training period to achieve a better energy optimization result, so providing an excellent starting point for the strategy by initializing behavioral strategies through supervised learning is believed to reduce the training cost of reinforcement learning strategies [124, 125]. In the energy management problem of MPS-EV, the introduction of expert knowledge and other policy parameters to initialize reinforcement learning behavioral policy has been widely used. Unlike the previous research that initiated the Q-learning policy with zero or random Q values, Xu et al. were inspired by the idea of the warm-start to initialize the Q-table values, giving higher initial values to the more efficient actions of the engine. The proposed method reduced the learning iterations of the algorithm compared to the standard Q-Learning [126]. Conventional Q-learning combines all states of the powertrain into the Q-table. In an FCEV-based EMS [127], the state-actions are optimized based on expertise. The state-action was restricted to a reasonable range of outputs for fuel cells and lithium batteries. This approach reduced the computational cost while improving the training effectiveness.

Related research [37, 38] used TL for knowledge transfer between hybrid vehicles with different structures to improve the efficiency of EMS development. Lian et al. studied knowledge transfer between four types of HEVs [37]. Firstly, the DDPG based on Toyota Prius was trained, the acquired deep neural network parameters were used to initialize the DNNs of the other models of EMS, and the DNNs initialized by TL were used for training. The results showed that the deep neural network initialized by TL improved convergence speed and fuel economy over the randomly initialized neural network. This result showed that reasonable initialization helps accelerate the development process of MPS-EV EMS.

## 4. Development Tendency of RL-EMS

This section presents trends that have not yet been widely used in the MPS-EV EMS field and explores the challenges and possibilities of applying these trends to RL-EMS in the future. The possible development direction of the algorithm in the future is also introduced.

*4.1 Trends in the application of RL algorithms on EMS*

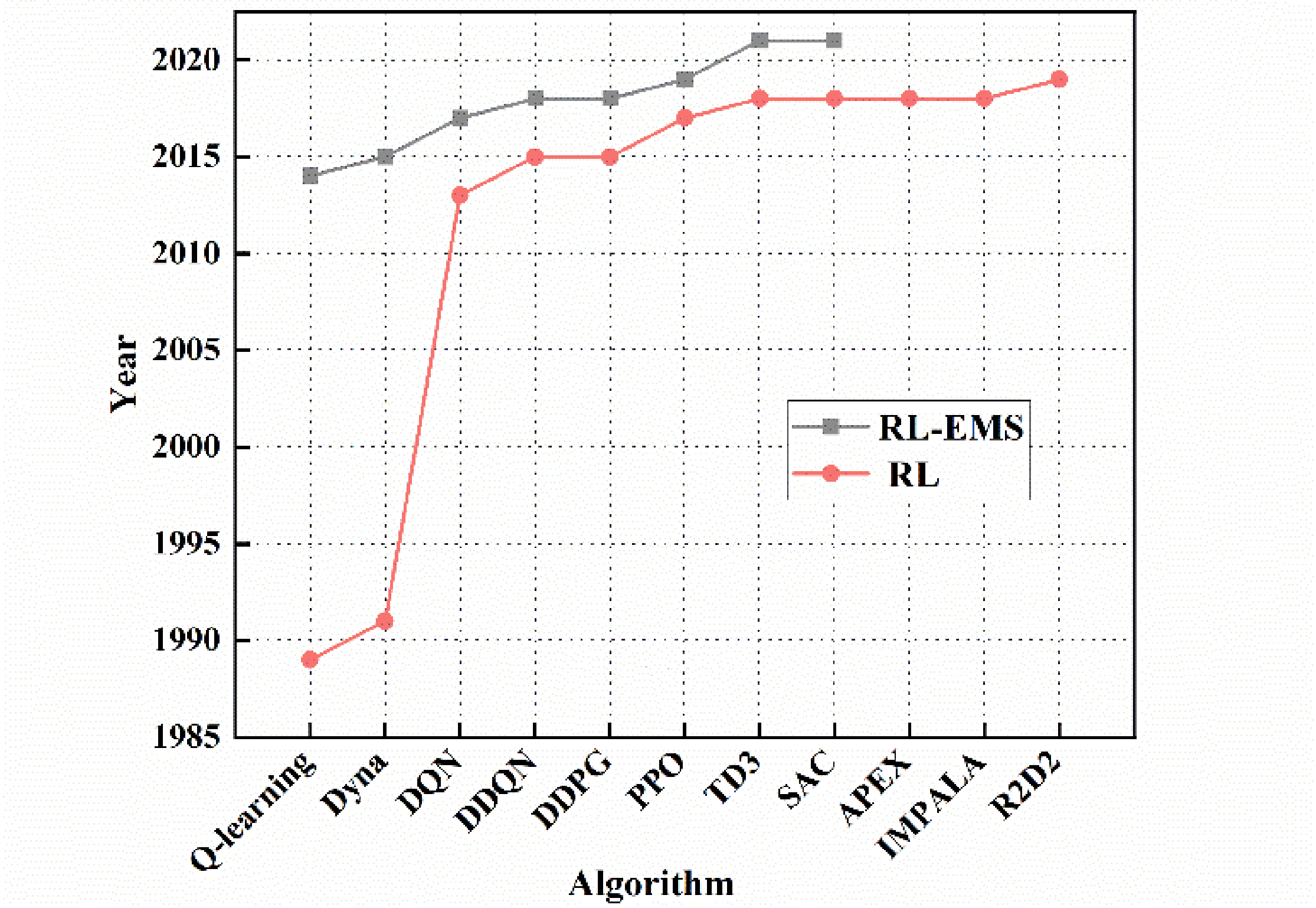

**Figure 16.** Time of RL algorithm proposal and early application in MPS-EV EMS research.

As shown in the figure, this paper lists the publishing time of some earlier research mentioned in this paper that applied reinforcement learning to MPS-EV energy management [36, 47, 61, 67, 70, 85, 92, 96]. In addition, the time when these algorithms were proposed is also presented. There is a gap between the proposal of the RL algorithms and its gradual application to the MPS-EV energy management problem, which can also be divided into three phases. Before 2013, reinforcement learning had not been applied to energy management. Early reinforcement learning algorithms like Dyna and Q-learning were proposed in the 20th century [60,128]. Despite their great significance in reinforcement learning, there was much room for improvement because they were relatively primitive. For a long time, the reinforcement learning algorithms did not attract the attention of researchers in the field of EMS until 2013, when Deep mind first proposed DQN. Reinforcement learning used deep neural networks to achieve control over a continuous state space, demonstrating its self-learning effect on Atari games [46], and an improved version, Double DQN was proposed in 2015 [129]. After entering the second phase, a series of deep reinforcement learning algorithms were proposed between 2013 and 2018. Due to the influence of the explosive development of reinforcement learning, there has been a proliferation of EMS research based on reinforcement learning. These algorithms have demonstrated the potential of RL-EMS, but at the same time, RL-EMS also has disadvantages such as poor robustness, hyperparameter sensitivity, and difficulty in deployment. With the advancement of communication technology, the expansion of application scenarios, and the complexity of control problems, AI development focuses on big data, big models, and application scenarios with high data flux. After 2018, with the iteration of reinforcement learning algorithms, some latest distributed offline reinforcement learning algorithms emerged with the technical potential to further enhance RL-EMS.

### *4.2 Differences in the development of RL-EMS and advanced RL algorithms*

#### *4.2.1 Introduction of some RL algorithms that have not yet been applied*

Although a series of advanced RL algorithms, such as SAC and TD3, have been applied, some advanced offline distributed algorithms have not been explored by relevant researchers in the energy management field for some reasons.

Distributed prioritized experience replay (APEX) [130] proposes a distributed architecture for large-scale deep reinforcement learning that enables learners to learn efficiently from orders of magnitude data. APEX allows distributed Actors to acquire experience data and initialize the priority level of experience for traditional DQN and DDPG. The learner updates network parameters based on experience in the database, passes the parameters to actors, and updates the priority level of the experience. Thus, the reinforcement learning network can be trained more efficiently. Studies have shown significant experimental results using this architecture on DQN and DDPG, improving the exploration efficiency with an increasing number of actors [130].

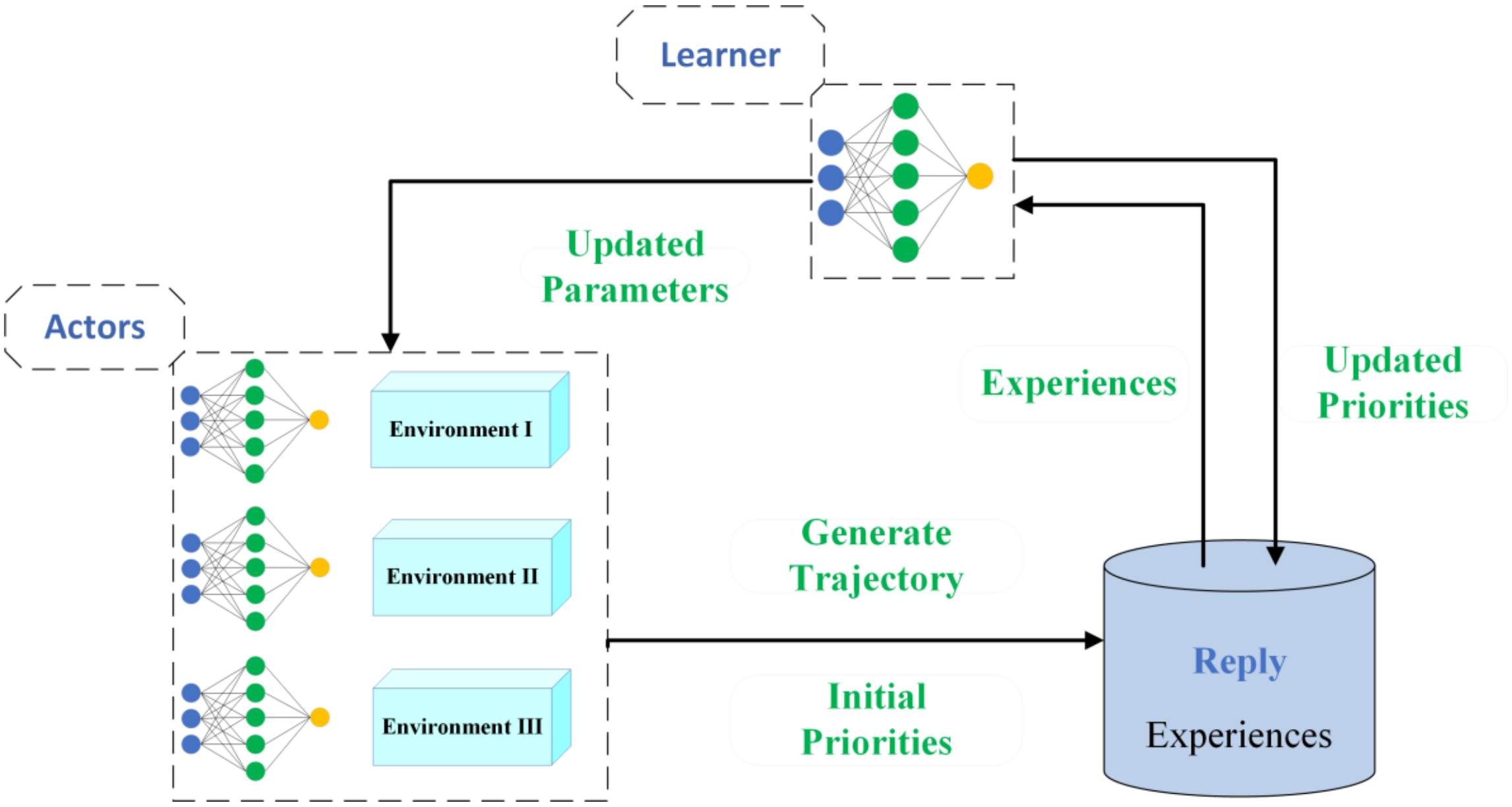

**Figure 17.** APEX algorithm framework.

Importance Weighted Actor-Learner Architecture (IMPALA) [131] proposes a new offline distributed AI with a classical actor-critic structure, where the learner uses the GPU and the actor uses the CPU. Each actor individually and periodically receives data from the learner to synchronize parameters and then collects data. Given that the experience replay mechanism consumes high memory and the learner's learning is batch-by-batch serial learning by mini-batch, which has a limitation on data throughput, the data collected by all the actors of IMPALA are stored in the data sampling queue instantly. The actors and the learner do not interfere with each other. The actor regularly synchronizes parameters from the learner with no need to calculate the gradient, and the data throughput is larger. Learner quantitatively learns to update parameters. The introduction of V-trace solves the problem that the policy when actors collect data uses is different from the policy when learner learns, which can accept a larger delay between them and facilitate large-scale distributed deployment of actors. IMPALA supports both single learner and multiple synchronous learners, and when the training scale increases, multiple learners with multiple actors can be considered. Each Learner only gets samples from its own actor to update. The learners exchange gradients and update network parameters periodically, and the actor updates parameters from any learner.

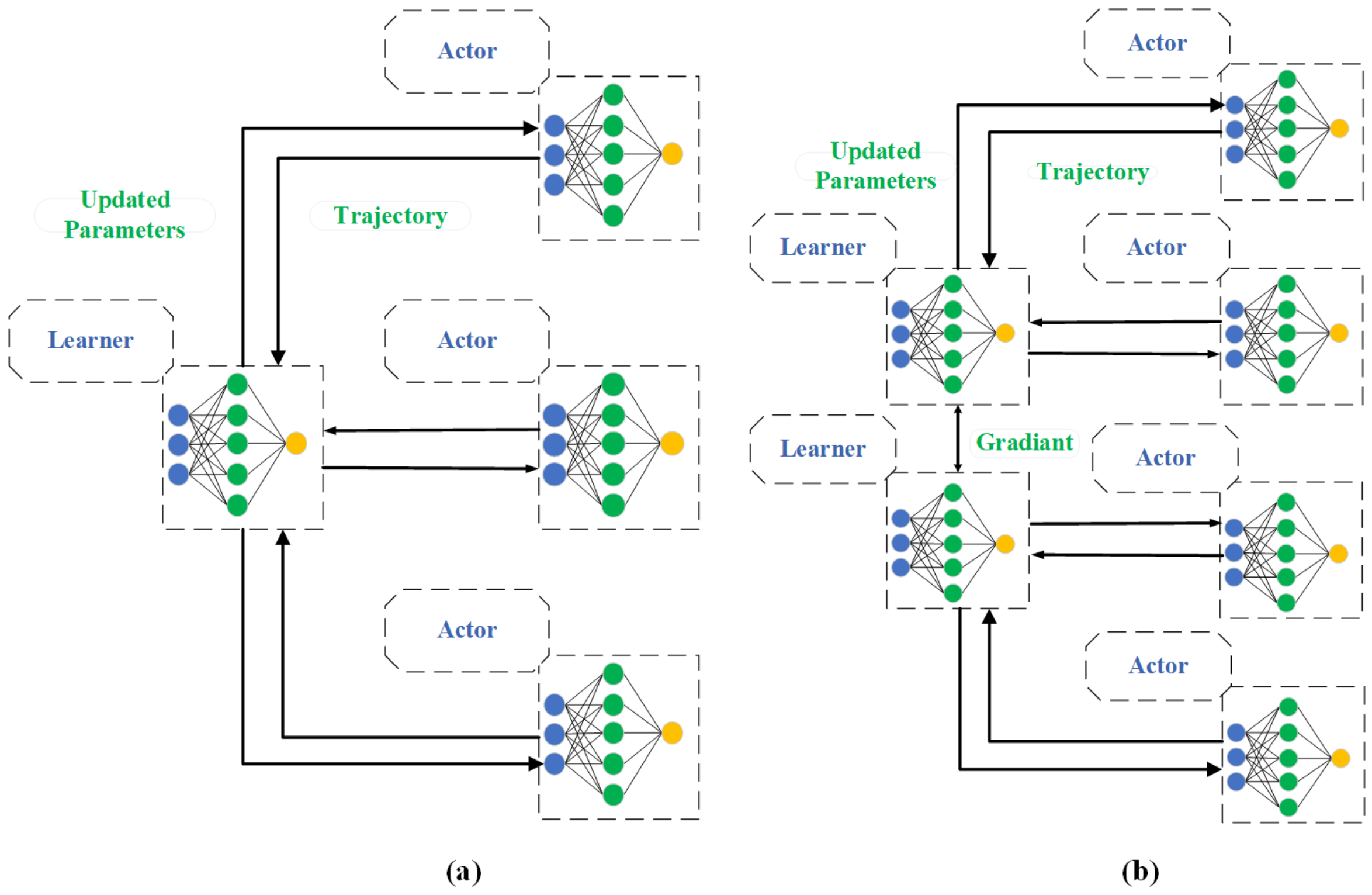

**Figure 18.** IMPALA algorithm framework. (a) Single Learner. (b) Multiple Synchronous Learners.

Recurrent Replay Distributed DQN (R2D2) [132] was proposed in 2019, and DeepMind introduced several new improvements to the previous year's APEX-DQN to produce an even better R2D2 algorithm. Its distributed training framework was used to train in the mode of single learner and 256actors, achieving high efficiency [132].

*4.2.1 The advantages of offline distributed reinforcement learning and the reasons why it has not been applied*

In reinforcement learning, the agent needs to generate a large amount of experience by interacting with the environment and learn online based on the experience to update its policy. Data is a crucial part of the training process for reinforcement learning, which can also be divided into online and offline learning according to the data source. The above three algorithms belong to offline reinforcement learning, in which the agent can learn directly from the data shared in the database without taking action. By offline reinforcement learning, the cost of interaction can be reduced.

Past RL-EMS research has mostly been limited to online integrated AI: single agent learns by its intersection with the environment, learning experiences are generated solely from its own vehicle model and driving cycles with a single source of data, slow data acquisition, and wasted computational resources. However, a broad trend in deep learning is to combine more computation with more powerful models and larger data sets [133]. All three newly developed algorithms are computationally solved at scale as offline distributed AI, sharing data through a central database, enhancing learning sample diversity, and reducing the cost of repeated computations. The use of distributed actors to acquire experience data and upload it to learners makes it possible for learners to learn, acquire experience, and improve strategies from a broader range of scenarios.

The large amount of data requires more workers to interact with the environment to enrich the experience and also requires a computing platform with high data throughput to support the transmission and calculation. Most of the current RL-EMS research uses Matlab + Simulink for simulation, with fixed vehicle models, repetitive training data, and learning experience from a single source. Such a simulation environment does not have the conditions for large-scale data parallelism. In contrast, some distributed AI studies can use up to 44 actors running on CPUs in the experiment to generate data at up to 19K frames per second [134], while the single learner running on a GPU receives a large number of experiences and provides different exploration policies for different actors, expands the diversity of experiences, and makes progress on high-dimensional control problems. An approach like Matlab + Simulink cannot be deployed in massive clusters of computers. At the same time, the decentralized authorization characteristic of Matlab makes it challenging to deploy and execute remotely, which leads to the inability to extend the research of the automobile control system to the current mainstream heterogeneous computing platform, thus making the latest research results of the general artificial intelligence, including deep reinforcement learning, unable to be conveniently applied.

### *4.3 Potential Research Directions*

#### *4.3.1 Algorithm improvement*

A) Security Exploration

RL explores the global optimum by trial-and-error search. Therefore, it must be shaped by a large amount of data interacting with its environment, but safety issues during execution are not necessarily avoided during trial-and-error [135], especially when the physical constraints of the vehicle cannot be exceeded when the properties of the environment change. In this case, learners should consider the consequences of their actions and explore safely in the real physical world. For the energy management problem of MPS-EV, the safety research of reinforcement learning solutions can be carried out in the following directions.

a) Action Pruning: The exploration of reinforcement learning is often based on the random selection of actions or sampling of policy distributions. Such actions are potentially dangerous in MPS-EV energy management problems, where excessive charging or discharging of the battery at low or high SOC states can easily reduce the battery life, and overcharging of the battery may even trigger abnormal battery operation [136]. Therefore, filtering or masking some of the control actions of the agent that may lead to dangerous states during training and deployment of reinforcement learning is an intuitive disaster avoidance method, which is often referred to as action pruning or action masking. Gao et al. et al. have shown that using action pruning methods in high-hazard action space problems can be effective in avoiding training disasters to enhance the training effectiveness of strategies [137]. Studying and developing action pruning methods for MPS-EV energy management problems would be an effective solution to limit the action space of reinforcement learning in the critical state of vehicle components.

b) Human Intervention: With the further development of V2X and MPS-EV configurations, states and action spaces in the future EMS may break through the current research architecture, resulting in that the current research and definitions of hazard exploration actions cannot be inherited and applied. The introduction of human decision-making experience in critical states, which is often considered safe and trustworthy, would help improve the safety exploration process of reinforcement learning in future research on MPS-EV energy management problems. Saunders et al. have shown that direct human intervention could reduce safety exploration disasters in reinforcement learning [138]. How to introduce and develop human interventions to improve the exploration process of RL-EMS to enhance the training effect of reinforcement learning will become a research direction of interest.

c) Recovery: In addition to the safety issues caused by stochastic exploration, the noise of state perception, unreasonable reward functions, and differences in deployment environments can lead to potential safety problems [139]. Assuming that security incidents cannot be avoided, how to quickly recover from catastrophic events will be the focus of future research. Yang et al. proposed a safe reinforcement learning framework that switches between a safe recovery strategy that prevents an agent from entering an unsafe state and a learner strategy that optimizes for task completion. Their experimental results showed that the safe recovery strategy effectively reduces the occurrence of safety accidents and improves the training effectiveness of reinforcement learning strategies [140]. Safe recovery strategies for reinforcement learning in MPS-EV energy management problems have not been investigated. The introduction of safe recovery strategies in the foreseeable future will greatly improve the safety of exploration of reinforcement learning in MPS-EV energy management problems.

B) Robustness improvement

In order to apply the RL algorithm to the energy management problem of MPS-EV more widely, the road environment and driving styles of each agent should be taken into consideration. Decisions made only for instantaneous speed and acceleration information cannot be extended to many different working conditions with different styles. In the past, RL-EMS handled a limited sample, which made it difficult to handle the diversity of real-world testing driving cycles. There are three research directions to be explored for the poor robustness of RL-EMS.

a) Real-time generation of training data: Although reinforcement learning algorithms can adapt to different driving cycles by self-learning effects through experience gained from interaction with the environment, iteration of the strategy requires computational costs and some time to adapt when facing real driving cycles that differ significantly from the training driving cycles. Therefore, the training data are required to be close to the real world and have diversity. Some research carried out training on fixed driving cycles, which lacks adaptability. To address this drawback, some researchers have enhanced the diversity of training data by introducing driving cycles with a large period collected from real roadways [141], generating real-time conditions based on simulation software to simulate different styles of road conditions for the vehicle model [117], adding noise to the original training data [36]. Markov chain model is used to generate driving cycles close to the latest historical driving data in real time for simulation [90]. There are gaps between the strategies generated based on simulation models and the real world, real-world working conditions data are more difficult to obtain, the virtual model cannot perfectly fit the actual power performance of the vehicle, and there are deficiencies in on-board computing system arithmetic power. These require the agent to generate sufficient experience to guide energy distribution based on diverse information. Research on improving RL-EMS robustness through real-time generated driving context information is emerging as a potential research hotspot.

b) Feature extraction for working conditions: RL agent makes decisions through state variables. The richness of training working conditions allows the agent to train on driving cycles with changing styles, but the state transfer probability matrix changes when the overall driving style is unknown and changes. It is difficult for an agent to plan close to the global optimum in multiple driving cycles based only on transient information such as speed and acceleration within a single time step. This requires the agent to extract the characteristics of entire working conditions and to sense the changes in overall working conditions in time. Guo et al. trained DDPG networks according to different speed intervals [80]; Lin et al. calculated the state transfer probability matrix in different time intervals during the driving time and measured the difference in their KL scatter as a basis for updating the strategy to achieve adaptability [122]. How to improve the sensing capability of RL-EMS agents and enhance the strategy robustness by introducing statistical and temporal features of the working conditions will be a potential future research direction.

c) Macroscopic traffic information sensing: With the development of V2X technology, vehicles can sense real-time traffic conditions, such as congestion and traffic light on surrounding and preceding roads. Weather conditions are also an important factor affecting driver behavior as well as fuel and power costs [142]. Much of the information provided by V2X can reflect the driving cycles the agent deals with to some extent and predict future travel information to inspire RL. Both speed prediction and lane density prediction based on macroscopic traffic information will contribute to the application of RL-EMS to the MPS-EV energy management problem.

### *4.3.2 Deployment solution*

The current development of AI tends to rely on big models and big data, and advanced AI solutions show significant high computing power demand characteristics. Therefore, future RL-EMS research will certainly face AI deployment challenges. Deployment on a single vehicle requires relevant deployment architecture tools. The following three main deployment tools are widely used for AI solutions: Open Visual Inference & Neural Network Optimization (OpenVino) for CPU devices; TensorRT for GPU devices; and MediaPipe, a deployment tool for edge devices.

TensorRT is a high-performance deep learning optimizer that provides low-latency, high-throughput deployments for deep learning applications. TensorRT can be used to compute acceleration for large-scale data centers, embedded platforms, or autopilot platforms. TensorRT is now capable of supporting TensorFlow, Caffe, Mxnet, Pytorch, and almost all other deep reinforcement learning frameworks. Combining TensorRT with high-performance GPUs for mobile devices enables fast and efficient deployment and computation in a wide range of industrial scenarios.

The OpenVINO tool suite is compatible with trained models from a variety of open source frameworks and has various capabilities for on-line deployment of algorithmic models, allowing pre-trained models to be quickly deployed on the CPU.

MediaPipe is a graphics-based cross-platform framework for building machine learning pipelines for multi-mode (video, audio, and sensor) applications. MediaPipe can run across platforms on mobile devices, workstations, and servers, and supports mobile GPU acceleration. MediaPipe can not only be deployed on the server side but also serve as an on-device machine learning inference framework on multiple mobile terminals and embedded platforms.

At present, RL-EMS based on deep reinforcement learning has achieved economy close to that of DP. However, in the future real vehicle deployment, no breakthrough has been made in how to deploy this effective AI scheme to vehicles without damage. Therefore, the combination of RL-EMS and AI deployment framework will become the focus of future research.

## **5. Conclusion**

This paper reviews the energy management strategies for multi-energy source vehicles based on reinforcement learning. The basics of reinforcement learning and multi-power source vehicles are introduced. The general process of RL-EMS is explained. An extensive review is focused on the application of various RL algorithms in the EMS field, and it can be seen that advanced RL algorithms can facilitate the performance of EMS. State sensing and control action schemes for RL-EMS are outlined. The setting of the reward function in RL-EMS is summarized. A description of the improvements to the RL-EMS training method due to vehicle characteristics and real driving cycles is presented. In this paper, we believe that one of the future research directions is to take advantage of the large-scale data utilization of offline distributed deep reinforcement learning to address the shortcomings of centralized reinforcement learning in the past. Accelerating the implementation of advanced algorithms with corresponding hardware and software for the real-time real-world deployment and enhancing the safety and robustness of the algorithms are necessary to realize the construction of energy management strategies in intelligent transportation environments.

## Author contributions

Conceptualization: J.H., Y.L. Writing of Original Draft: Y.L. Revision: L.C., Y.Z. Supervision: Y.Z., J.H., J.J., J.L, Z.H.

## Declaration interests

The authors declare no competing interests.